\documentclass{article}

\usepackage{microtype}
\usepackage{graphicx}
\usepackage{subcaption}
\usepackage{booktabs} 
\usepackage[utf8]{inputenc}
\usepackage{siunitx}  
\usepackage{multirow} 
\usepackage{enumitem}
\DeclareMathAlphabet{\mathmybb}{U}{bbold}{m}{n}

\usepackage{hyperref}

\usepackage[preprint]{icml2026}

\usepackage{amsmath}
\usepackage{amssymb}
\usepackage{mathtools}
\usepackage{amsthm}

\usepackage[capitalize,noabbrev]{cleveref}

\theoremstyle{plain}

\theoremstyle{definition}

\theoremstyle{remark}

\usepackage[textsize=tiny]{todonotes}

\icmltitlerunning{Excitation: Momentum For Experts}

\begin{document}

\twocolumn[
  \icmltitle{Excitation: Momentum For Experts}



  \icmlsetsymbol{equal}{*}

  \begin{icmlauthorlist}
      \icmlauthor{Sagi Shaier}{aar}
  \end{icmlauthorlist}

    \icmlaffiliation{aar}{Aleph Alpha Research, Heidelberg, Germany}

  \icmlcorrespondingauthor{Sagi Shaier}{sagi.shaier@aleph-alpha-research.com}

  \icmlkeywords{Machine Learning, ICML}

  \vskip 0.3in
]



\printAffiliationsAndNotice{}  

\begin{abstract}
We propose \textsc{Excitation}, a novel optimization framework designed to accelerate learning in sparse architectures such as Mixture-of-Experts (MoEs). Unlike traditional optimizers that treat all parameters uniformly, \textsc{Excitation} dynamically modulates updates using batch-level expert utilization. It introduces a competitive update dynamic that amplifies updates to highly-utilized experts and can selectively suppress low-utilization ones, effectively sharpening routing specialization. Notably, we identify a phenomenon of ``structural confusion'' in deep MoEs, where standard optimizers fail to establish functional signal paths; \textsc{Excitation} acts as a specialization catalyst, ``rescuing'' these models and enabling stable training where baselines remain trapped. \textsc{Excitation} is optimizer-, domain-, and model-agnostic, requires minimal integration effort, and introduces neither additional per-parameter optimizer state nor learnable parameters, making it highly viable for memory-constrained settings. Across language and vision tasks, \textsc{Excitation} consistently improves convergence speed and final performance in MoE models, indicating that active update modulation is a key mechanism for effective conditional computation.

\end{abstract}

\section{Introduction}

\begin{figure}[!t] 
  \centering
  \includegraphics[width=0.9\linewidth]
  {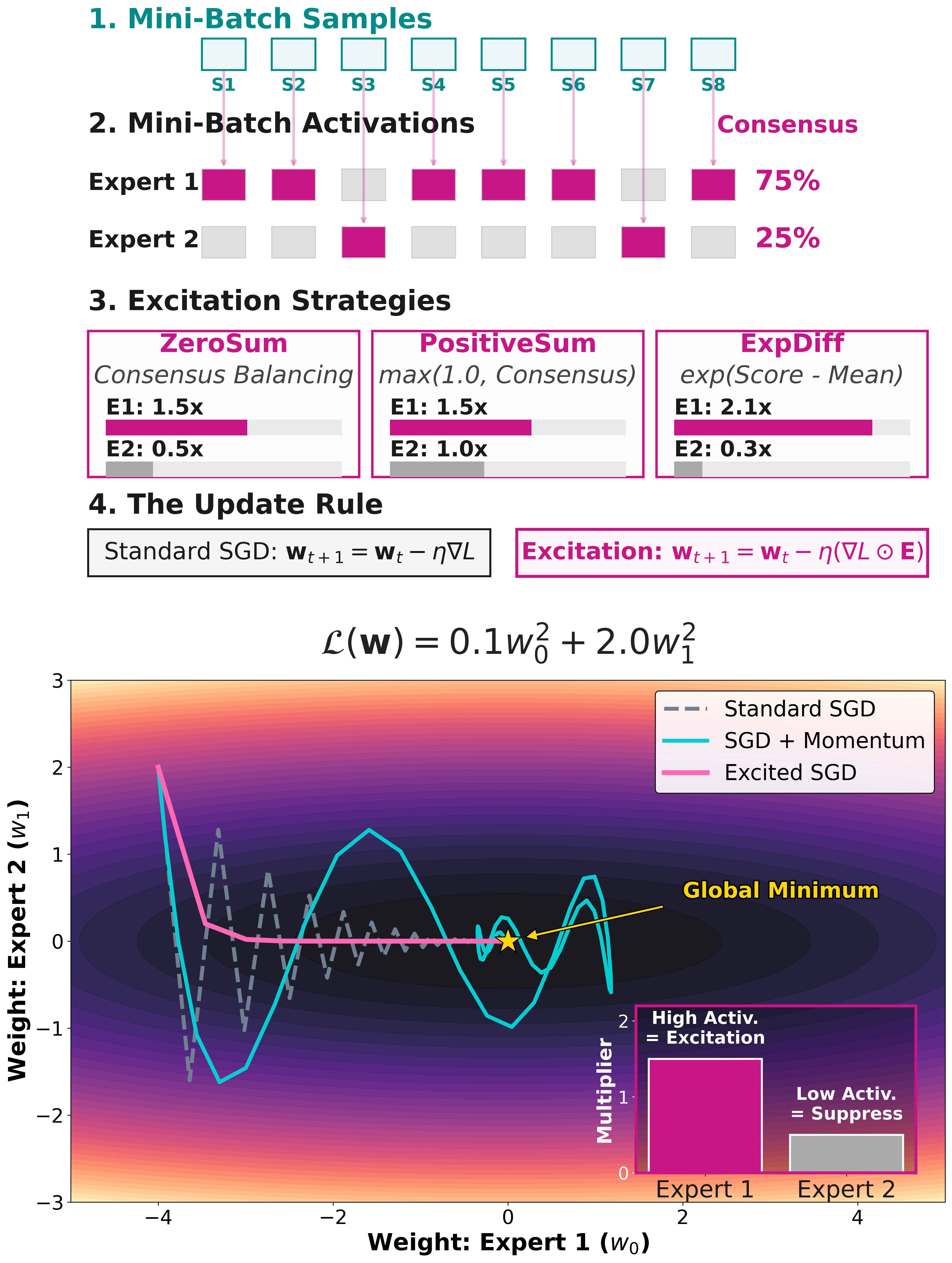}
  \caption{\small \textbf{Overview of \textsc{Excitation}.} \textit{Top:} Batch-level expert activations drive competitive update modulation—three strategies are available (here: ZeroSum with 1.5×/0.5× multipliers). \textit{Bottom:} Optimization trajectory on a 2D toy problem using ZeroSum. Excited SGD (pink) converges faster by amplifying high-consensus expert updates (Expert 1, $w_0$) while suppressing low-utilization ones (Expert 2, $w_1$), outperforming standard SGD (gray)
  and SGD+Momentum (cyan).}
  \label{fig:figure1}
\end{figure}

Mixture-of-Experts (MoE) architectures have emerged as a dominant scaling paradigm, enabling conditional computation where only a sparse subset of parameters is activated per input \cite{bengio2013estimatingpropagatinggradientsstochastic, shazeer2017outrageouslylargeneuralnetworks, bengio2016conditionalcomputationneuralnetworks, fedus2022switchtransformersscalingtrillion}. By decoupling total parameter count from per-token cost, MoEs allow models to scale to trillions of parameters while remaining computationally efficient—a strategy now ubiquitous in state-of-the-art language and vision systems \cite{wu2024deepseekvl2mixtureofexpertsvisionlanguagemodels, deepseekai2025deepseekv3technicalreport, Cai_2025, jiang2024mixtralexperts}.

Despite their promise, MoEs present unique optimization challenges. Unlike dense networks, MoE updates are gated by routing decisions, creating a feedback loop: rarely selected experts remain under-trained while frequently selected ones rapidly specialize. This often culminates in \textbf{expert imbalance and specialization collapse}, where a small subset of experts dominates computation while others remain redundant \cite{shazeer2017outrageouslylargeneuralnetworks, fedus2022switchtransformersscalingtrillion}. Such mismatches degrade capacity utilization and training stability \cite{wang2024auxiliarylossfreeloadbalancingstrategy, clark2022unifiedscalinglawsrouted}.

Prior work addresses these dynamics via routing regularization \cite{zoph2022stmoedesigningstabletransferable} or load-balancing losses \cite{shazeer2017outrageouslylargeneuralnetworks}. However, these interventions occur at the architectural or objective level, leaving the update rule \textit{agnostic to routing assignments}. Because standard adaptive optimizers \cite{kingma2017adammethodstochasticoptimization} treat all parameters identically, rarely activated experts receive sporadic updates susceptible to small-sample noise, while frequently selected, high-utilization experts are not explicitly prioritized.

We argue that this \textbf{routing-blind optimization} prevents the framework from effectively catalyzing functional differentiation. In deep sparse networks, this manifests as \textbf{``structural confusion,''} where standard update rules fail to establish stable signal paths through conditional gates. By ignoring the stochasticity of expert selection, conventional optimizers struggle to reinforce the precise, high-confidence routing required for effective conditional computation.

To address this gap, we propose \textsc{Excitation} (Fig.~\ref{fig:figure1}), an activation-aware optimization framework. \textsc{Excitation} dynamically \textbf{modulates updates based on batch-level expert utilization}, reinforcing high-consensus routing decisions. This encourages functional specialization and reduces routing entropy without requiring architectural modifications or auxiliary objectives. Unlike prior adaptive methods, \textsc{Excitation} requires no additional optimizer state nor learnable parameters, and is \textbf{compatible with all optimizers}.

Our evaluation across diverse MoE architectures in vision and language demonstrates that \textsc{Excitation} consistently improves convergence speed and final performance. These results establish active update modulation as a robust, domain-agnostic mechanism for stabilizing and accelerating the training of sparse neural networks.

Our core contributions are:
\begin{itemize}[leftmargin=*, nosep]
    \item \textbf{The \textsc{Excitation} Framework:} We introduce a novel optimization framework that modulates updates based on batch-level expert utilization. It is model-agnostic, requires no additional optimizer state, and imposes negligible computational overhead.    
    \item \textbf{Mechanism of Action:} We show that \textsc{Excitation} forces a transition from noisy, redundant routing to deterministic specialization. We identify the phenomenon of \textit{structural confusion} and demonstrate that our framework enables ``structural rescue'' in deep regimes where standard optimizers fail to converge.
    \item \textbf{Empirical Validation:} Extensive evaluation across Vision and Language MoEs shows consistent gains in convergence speed and final  accuracy/perplexity.
    \item \textbf{Robustness \& Scaling:} Sensitivity analysis confirms robustness to hyperparameter tuning and 
    that performance increase monotonically with sparsity—reaching $+5.57\%$ at 90\% sparsity—with $<0.1\%$ overhead at scale.    
\end{itemize}

\section{Related Work}

\textbf{Scaling via Conditional Computation}
MoEs enable massive scaling by decoupling total parameter count from per-token computational costs \cite{shazeer2017outrageouslylargeneuralnetworks, lepikhin2020gshardscalinggiantmodels, jiang2024mixtralexperts}. Despite their efficiency, these models often suffer from \textit{expert collapse} or high router entropy \cite{chi2022representationcollapsesparsemixture, shazeer2017outrageouslylargeneuralnetworks, wang2024auxiliarylossfreeloadbalancingstrategy, shaier2025expertsgalaxiesconditionallyoverlappingexperts, lewis2021baselayerssimplifyingtraining, thiombiano2025moxemixturexlstmexperts, do2023hyperrouterefficienttraininginference}, where routing remains stochastic and experts fail to develop distinct functional boundaries. Prior work addresses this via architectural constraints—such as Expert-Choice routing \cite{zhou2022mixtureofexpertsexpertchoicerouting} or Top-K gating \cite{nguyen2024statisticalperspectivetopksparse}—yet we posit the root instability often lies in optimization dynamics. Similar phenomena appear in dynamic sparse training, where inadequate gradient flow prevents stable computational paths \cite{evci2021rigginglotterymakingtickets, Mocanu_2018, frankle2019lotterytickethypothesisfinding}. Unlike methods modifying connectivity or routing logic, \textsc{Excitation} isolates the failure mode at the update level, demonstrating that utilization-aware modulation can ``rescue'' configurations that would otherwise structurally diverge.

\textbf{Routing-Blind Adaptive Optimization}
Modern adaptive optimizers including Adam \cite{kingma2017adammethodstochasticoptimization}, AdamW \cite{loshchilov2019decoupledweightdecayregularization}, and LAMB \cite{you2020largebatchoptimizationdeep} scale updates based on gradient moments but remain \textit{oblivious} to routing mechanisms. In MoEs, routing explicitly controls parameter participation, yet this context is not exposed to the optimizer. Update magnitudes are thus determined independently of activation frequency, creating a fundamental misalignment: underutilized experts receive uncalibrated updates while high-consensus experts lack explicit reinforcement to solidify functional roles. While gradient clipping \cite{zhang2020gradientclippingacceleratestraining} or LR warm-up \cite{kalra2024warmuplearningrateunderlying} provide global stability, they do not resolve this \textbf{routing-blindness}. Existing work on parameter-wise scaling \cite{you2017largebatchtrainingconvolutional} or memory-efficient adaptivity \cite{shazeer2018adafactoradaptivelearningrates} treats parameter participation as dense, ignoring the unique stochasticity of conditional computation. \textsc{Excitation} bridges this gap by incorporating batch-level expert utilization directly into the update rule.

\textbf{Utilization-Aware Learning Dynamics}
The prevailing strategy for managing expert imbalance relies on auxiliary objectives like load-balancing or router $z$-losses \cite{shazeer2017outrageouslylargeneuralnetworks, fedus2022switchtransformersscalingtrillion, zoph2022stmoedesigningstabletransferable}. These penalize \textit{symptoms} of imbalance by regularizing router outputs but ignore \textbf{parameter-level learning progress}. \textsc{Excitation} offers a complementary optimizer-level intervention: by amplifying high-consensus updates, it aligns optimization trajectories with router specialization. This addresses credit assignment challenges in conditional computation \cite{bengio2016conditionalcomputationneuralnetworks, bengio2013estimatingpropagatinggradientsstochastic, graves2017adaptivecomputationtimerecurrent}, where gradient signals are sparse, delayed, or stochastic \cite{kirsch2018modularnetworkslearningdecompose}. Rather than relaxing \textbf{routing discreteness} or modifying gradient estimators, we reshape the update manifold using realized utilization, catalyzing functional differentiation without additional structural constraints.

\section{The \textsc{Excitation} Framework}
Let $\theta \in \mathbb{R}^d$ be partitioned into $K$ disjoint parameter sets $\{\theta^{(1)}, \dots, \theta^{(K)}\}$ corresponding to experts. In sparse architectures, only a subset of these experts is activated for any given input. In standard first-order optimization, the update for expert $k$ at step $t$ is typically $\theta_{t+1}^{(k)} = \theta_t^{(k)} + \Delta \theta_t^{(k)}$.

\subsection{Activation-Aware Modulation}
We propose \textsc{Excitation}, a framework that dynamically re-scales updates based on expert utilization within a batch. We define the utilization vector $u \in [0, 1]^K$ as:
\begin{equation}
    u_k = \frac{1}{|\mathcal{B}|} \sum_{x \in \mathcal{B}} \mathmybb{1}[k \in \mathcal{A}(x)]
\end{equation}
where $\mathcal{A}(x)$ denotes the set of active experts for input $x$. The modulated \textsc{Excitation} update is then:
\begin{equation}
    \Delta \theta_t^{(k), Exc} = \Phi(u_k, \gamma) \cdot \Delta \theta_t^{(k)}
\end{equation}
where $\Phi(\cdot)$ is an \textsc{Excitation} function and $\gamma$ is a non-learnable hyperparameter controlling the intensity of the excitation.

\paragraph{Optimizer-Agnostic Implementation.} 
\textsc{Excitation} operates on the resultant parameter change rather than internal optimizer logic. For any base optimizer $\mathcal{O}$ (e.g., Adam, SGD), the \textit{Excited} update is computed as:
\begin{align}
    \delta_t^{(k)} &= \mathcal{O}(\theta_t, \nabla_{\theta} \mathcal{L}_t)^{(k)} - \theta_t^{(k)} \\
    \theta_{t+1}^{(k)} &= \theta_t^{(k)} + \Phi(u_k, \gamma) \cdot \delta_t^{(k)}
\end{align}
This ensures \textsc{Excitation} requires no additional per-parameter optimizer state, maintaining memory and throughput efficiency (cf. Appendix~\ref{comp_efficiency}). For example, the \textbf{Excited Adam} update becomes:
\begin{equation}
    \theta_{t+1}^{(k)} = \theta_t^{(k)} + \Phi(u_k, \gamma) \cdot \left( -\eta \frac{\hat{m}_t^{(k)}}{\sqrt{\hat{v}_t^{(k)}} + \epsilon} \right)
\end{equation}
By amplifying high-utilization updates, we prioritize experts with high batch consensus while optionally suppressing parameters with stale or noisy gradient information.

\subsection{Excitation Functions}
We propose three primary strategies and three control variants to isolate the mechanism of improvement.

\subsubsection{Core Formulations}
These functions are designed to sharpen functional specialization by rewarding high-consensus experts.

\begin{itemize}[leftmargin=*, nosep]
    \item \textbf{Zero-Sum (ZS) Excitation:} This formulation implements the principle of \textit{gradient energy preservation}. By normalizing the power-scaled utilization, we ensure the average update magnitude remains constant:
    \begin{equation} \Phi_{ZS}(u_k, \gamma) = \frac{u_k^\gamma}{\mathbb{E}[u^\gamma]} \end{equation}
    where $\gamma$ is the power-law hyperparameter. This forces a competitive dynamic: to \textit{excite} active experts, the optimizer must proportionally \textit{suppress} inactive ones.
    \item \textbf{Positive-Sum (PS) Excitation:} A non-competitive variant that prevents the suppression of under-utilized experts:
    \begin{equation}
        \Phi_{PS}(u_k, \gamma) = \max(1.0, \Phi_{ZS}(u_k, \gamma))
    \end{equation}
    This formulation tests whether performance gains require explicit suppression of \textit{neglected} experts or if amplifying \textit{popular} ones is sufficient.
    \item \textbf{Exponential-Diff (ExpDiff):} An aggressive modulation that scales updates based on absolute utilization to force rapid functional boundaries. We define this as a normalized exponential:
    \begin{equation} 
        \Phi_{ED}(u_k, \gamma) = \frac{\exp(\gamma u_k)}{\mathbb{E}[\exp(\gamma u)]} 
    \end{equation}
    While Zero-Sum scales updates based on the \textit{ratio} of utilization to the mean, ExpDiff scales based on the \textit{exponential distance}. This ensures that even marginal leads in activation result in significant divergence in gradient energy, effectively forcing the model to solidify expert boundaries more aggressively.   
\end{itemize}

\subsubsection{Control and Ablation Variants}
To verify that \textsc{Excitation} benefits from targeted update modulation rather than global effects, we define:

\begin{itemize}[leftmargin=*, nosep]
    \item \textbf{Global-Exp (Energy Control):} To isolate the effect of spatial targeting from total gradient magnitude, we apply a uniform boost $\Phi_{GE}$ to all experts:
    \begin{equation} \Phi_{GE} = \mathbb{E} \left[ \exp \left( \gamma \cdot \text{ReLU}(u - \mathbb{E}[u]) \right) \right] \end{equation}
     This determines if improvements stem from targeted specialization or from a higher effective learning rate.
    
    \item \textbf{Random-Boost (Spatial Control):} Computes $\Phi_{ZS}$ but randomly permutes the coefficients across experts. This tests if the \textit{alignment} between magnitude and activation is the driver of convergence, rather than the mere existence of a non-uniform update distribution.

    \item \textbf{Inverted (Contrarian Control):} Boosts low-utilization experts and suppresses high-utilization ones:
    \begin{equation} \Phi_{Inv}(u_k, \gamma) = \frac{u_k^{-\gamma}}{\mathbb{E}[u^{-\gamma}]} \end{equation}
    This pathological test checks if prioritizing neglected experts disrupts functional specialization.
\end{itemize}

\section{Experiments}
\label{sec:experiments}
To ensure reproducibility, all experiments utilized a \textit{strictly synchronized seeding protocol} where model initialization, data loading, and stochastic sampling were kept identical between baselines and \textsc{Excitation} variants. Performance deltas are thus attributable solely to the framework rather than stochastic variance. Except for the large-scale GPT-MoE sweeps—which were limited to single-seed runs due to computational constraints—all experiments were conducted across three independent random seeds.

All experiments utilize standardized hyperparameters widely adopted in the literature~\cite{dosovitskiy2021imageworth16x16words, brown2020languagemodelsfewshotlearners, liu2021learningturningneuralarchitecture, srećković2025batchsizeproblemrevisiting, zhang2025adamminiusefewerlearning}.
This configuration reflects a realistic training regime, particularly for large-scale models where exhaustive hyperparameter sweeps are computationally prohibitive. By maintaining fixed configurations across all methods, we ensure a fair comparison that evaluates the ``out-of-the-box'' efficacy of our framework. We further investigate the robustness of these choices via an extensive sensitivity analysis in Section~\ref{sensitivity_sec}.

\subsection{Foundational Benchmark: Sparse Structural Convergence}

\begin{figure}[]
  \centering
  \includegraphics[width=0.95\linewidth]{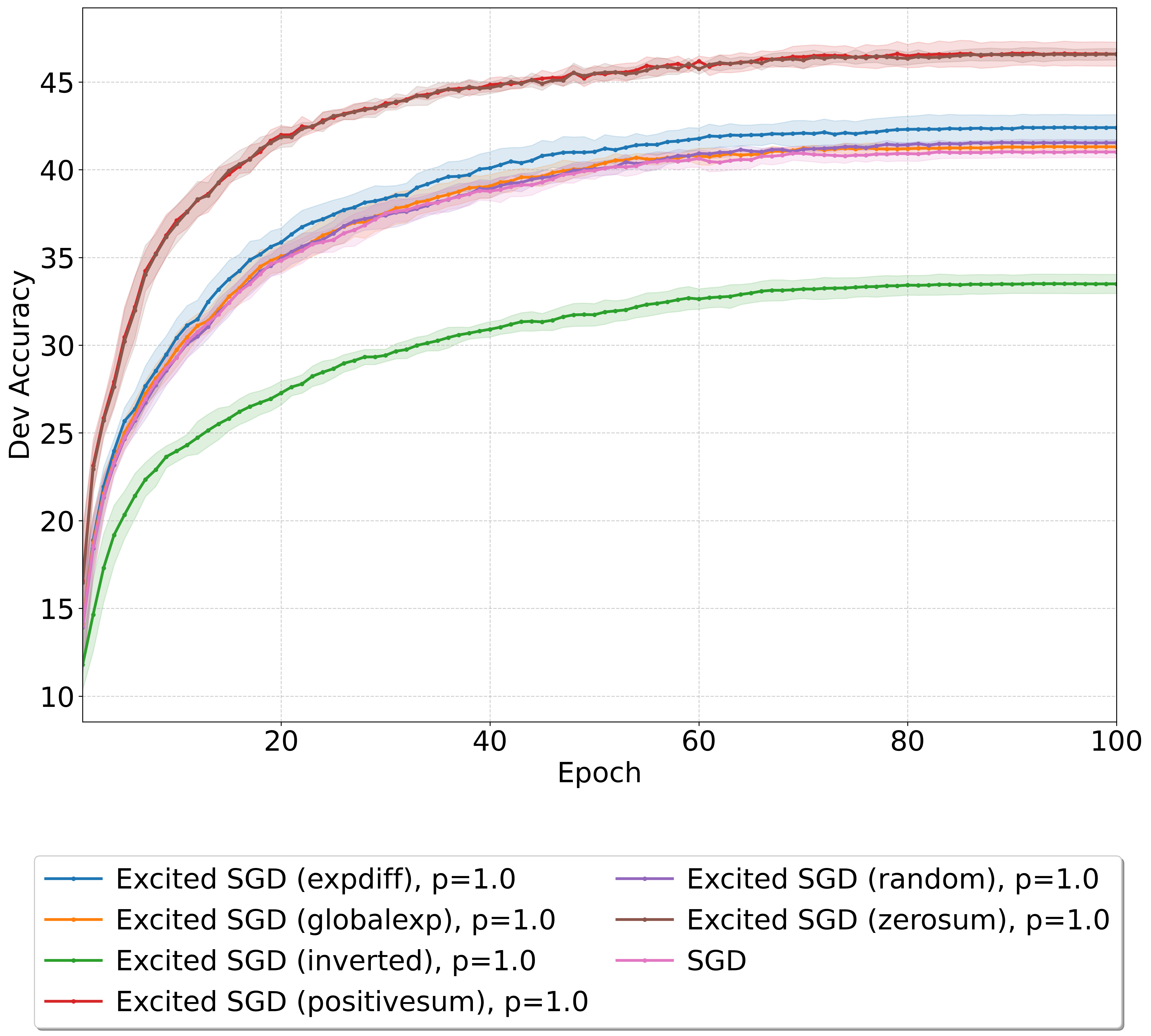}
  \caption{\textbf{CIFAR-10 Foundation Results.} Targeted mechanisms ($\Phi_{ZS}, \Phi_{PS}$) significantly increase accuracy. The failure of \textbf{Inverted}, \textbf{Random}, and \textbf{Global-Exp} variants demonstrates that success is driven by spatial reinforcement of expert specialization rather than simple learning rate scaling.}
  \label{fig:cifar10_foundation_sgd}
\end{figure}


We evaluate \textsc{Excitation} on CIFAR-10, CIFAR-100, and SVHN using a 4-layer MLP (128 neurons per layer), where in each layer only the Top-$K$ neuron activations are retained, corresponding to 90\% sparsity. This ``Micro-MoE'' setup isolates expert competition by treating individual neurons as experts. As shown in Fig.~\ref{fig:cifar10_foundation_sgd} (and App.~\ref{sgd_svhn_cifar100}), targeted variants ($\Phi_{ZS}, \Phi_{PS}, \Phi_{ED}$) deliver substantial gains in convergence speed and final accuracy.

\paragraph{Isolating the Targeting Effect}
To validate that success is driven by precise spatial targeting rather than simple global learning rate scaling, we analyze three control groups:

\begin{itemize}[leftmargin=*, nosep]
\item \textbf{Targeted allocation, not global energy, drives performance.} The \textbf{Global-Exp} control applies a uniform scalar boost to all parameters, effectively increasing the global learning rate. It yields only marginal improvements compared to $\Phi_{ZS}$ and $\Phi_{PS}$, confirming that success stems from directing gradient energy to active experts rather than a higher effective learning rate.
\item \textbf{Alignment between utilization and magnitude is the fundamental driver of performance.} The \textbf{Zero-Sum} ($\Phi_{ZS}$) formulation enforces update preservation ($\mathbb{E}[\Phi_{ZS}] = 1$), keeping the average update magnitude identical to the baseline. Its superiority over \textbf{Random-Boost} (shuffled spatial assignments) proves that the specific alignment between utilization consensus and update magnitude
drives the observed gains.
\item \textbf{Reinforcing established expert roles is essential for specialization.} The \textbf{Inverted} control, which amplifies low-utilization experts, significantly underperforms the baseline. This confirms that stable sparse training requires the reinforcement of high-consensus routing paths rather than the prioritization of neglected ones.
\end{itemize}


\subsection{Generalization Across Optimizers}
\begin{table}[]
\centering
\caption{Comparison of Base and Excited optimizers on CIFAR-10 (Top-$K$ MLP, $d=128, L=4$, sparsity $0.9$). $\Delta$ denotes absolute percentage improvement over the corresponding vanilla baseline.}
\label{tab:optimizer-comparison}
\small
\begin{tabular}{c c c c}
\toprule
\textbf{Base} & \textbf{Optimizer} & \textbf{Dev Accuracy (\%)} & \textbf{$\Delta$} \\
\midrule
\multirow{3}{*}{SGD} 
 & Vanilla & 41.01 $\pm$ 0.33 & --- \\
 & Excited (zs) & 46.58 $\pm$ 0.32 & \color{teal}{+5.57} \\
 & \textbf{Excited (ps)} & \textbf{46.59 $\pm$ 0.68} & \color{teal}{\textbf{+5.58}} \\
\midrule
\multirow{3}{*}{\shortstack{SGD\\(mom=0.9)}} 
 & Vanilla & 50.65 $\pm$ 0.20 & --- \\
 & Excited (zs) & 51.75 $\pm$ 0.24 & \color{teal}{+1.10} \\
 & \textbf{Excited (ps)} & \textbf{51.96 $\pm$ 0.30} & \color{teal}{\textbf{+1.31}} \\
\midrule
\multirow{3}{*}{Adam} 
 & Vanilla & 48.47 $\pm$ 0.23 & --- \\
 & Excited (zs) & 49.81 $\pm$ 0.29 & \color{teal}{+1.34} \\
 & \textbf{Excited (ps)} & \textbf{50.55 $\pm$ 0.24} & \color{teal}{\textbf{+2.08}} \\
\midrule
\multirow{3}{*}{AdamW} 
 & Vanilla & 48.48 $\pm$ 0.06 & --- \\
 & Excited (zs) & 50.23 $\pm$ 0.34 & \color{teal}{+1.75} \\
 & \textbf{Excited (ps)} & \textbf{50.67 $\pm$ 0.20} & \color{teal}{\textbf{+2.19}} \\
\midrule
\multirow{3}{*}{RMSprop} 
 & Vanilla & 48.47 $\pm$ 0.42 & --- \\
 & Excited (zs) & 50.20 $\pm$ 0.28 & \color{teal}{+1.73} \\
 & \textbf{Excited (ps)} & \textbf{50.66 $\pm$ 0.33} & \color{teal}{\textbf{+2.19}} \\
\midrule
\multirow{3}{*}{Adagrad} 
 & Vanilla & 32.17 $\pm$ 0.36 & --- \\
 & Excited (zs) & 35.35 $\pm$ 0.31 & \color{teal}{+3.18} \\
 & \textbf{Excited (ps)} & \textbf{35.45 $\pm$ 0.77} & \color{teal}{\textbf{+3.28}} \\
\bottomrule
\end{tabular}
\end{table}

\textsc{Excitation} is modular and agnostic to the underlying optimization algorithm. We evaluate its portability across a diverse suite of optimizers, including Adam, AdamW, SGD with Momentum, RMSprop, and Adagrad (Table \ref{tab:optimizer-comparison}). 
\textsc{Excitation} yields consistent improvements across all base methods: adaptive optimizers (Adam/AdamW) improve by approximately 2.2\%, while SGD and Adagrad exhibit gains exceeding 3.2\% and 5.5\%, respectively. These results indicate that \textit{reinforcing expert specialization is a complementary mechanism that operates independently of specific moment estimation or weight decay strategies}. Notably, the Positive-Sum variant ($\Phi_{PS}$) generally outperforms the Zero-Sum variant ($\Phi_{ZS}$), suggesting that while specialization is essential, softer constraints preserve exploratory capacity more effectively than strict zero-sum competition. 
Despite these 
differences, all \textsc{Excitation} variants maintain a robust efficiency advantage over vanilla baselines throughout training (cf. Appendix \ref{different_opt_training_curves}).

\subsection{Rescuing Deep Networks}
\label{deep_networks}

\begin{figure}[]
     \centering
     \begin{subfigure}[b]{0.48\linewidth}
          \centering
          \includegraphics[width=\linewidth]{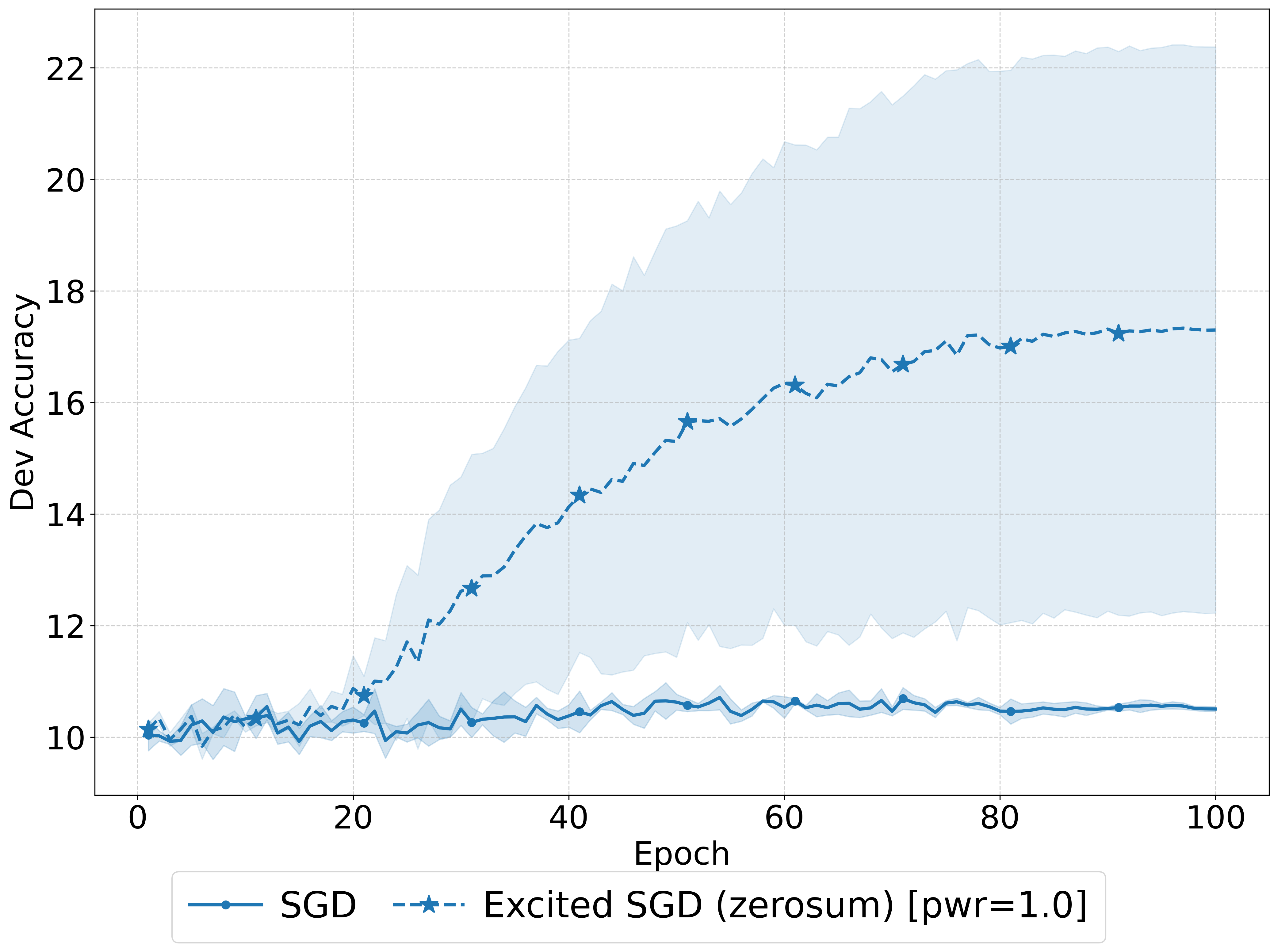}
          \caption{10-Layer SGD}
          \label{fig:10_layers_sgd}
     \end{subfigure}
     \hfill
     \begin{subfigure}[b]{0.48\linewidth}
          \centering
          \includegraphics[width=\linewidth]{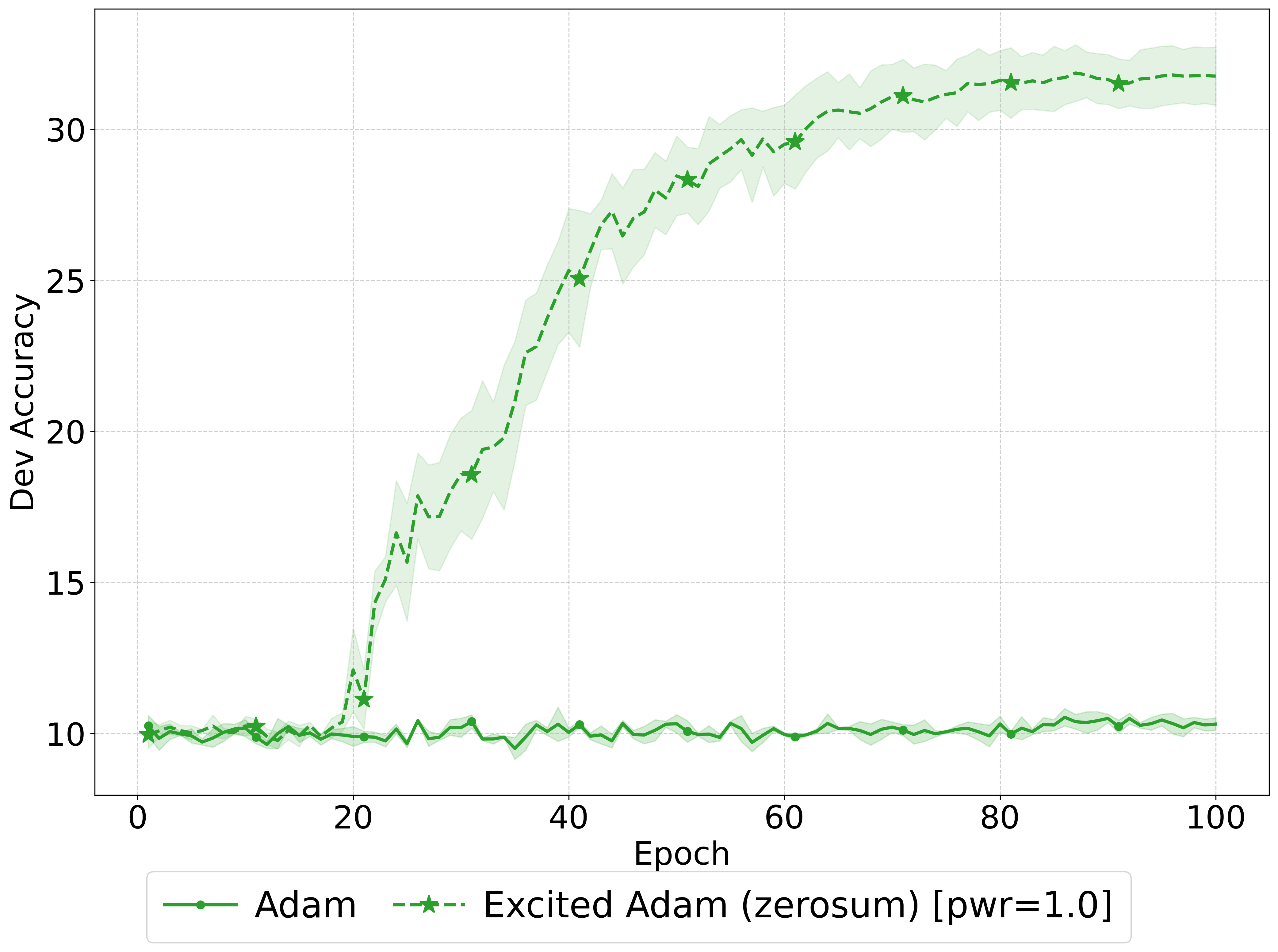}
          \caption{20-Layer Adam}
          \label{fig:20_layers_adam}
     \end{subfigure}
     \caption{\textbf{Rescuing deep sparse models from training collapse.} Standard optimizers remain trapped at near-random accuracy due to poor routing in deep layers. \textsc{Excitation} overcomes this bottleneck, enabling both SGD and Adam to ``escape'' the random-guessing regime and initiate meaningful specialization.}
     \label{fig:combined_escape_analysis}
\end{figure}

While skip connections mitigate instability, they alter the routing dynamics' signal-to-noise ratio. We evaluate \textsc{Excitation} in architectures without residuals to observe the ``raw'' formation of functional paths. In these regimes, we identify \textit{structural confusion}, where routing fails to establish viable signal paths, rendering deep models effectively untrainable. As shown in Fig.~\ref{fig:combined_escape_analysis}, baseline SGD and Adam remain trapped at random performance, 
whereas \textsc{Excitation} ``rescues'' these configurations by reinforcing early routing to prevent signal dissipation. In the 10-layer SGD (Fig.~\ref{fig:10_layers_sgd}) and 20-layer Adam (Fig.~\ref{fig:20_layers_adam}) settings, \textsc{Excitation} reaches 17\% and 33\% accuracy, respectively. These results demonstrate that \textsc{Excitation} can substitute for architectural scaffolding, enabling trainability in deep, conditional architectures where standard update rules fail to penetrate the sparse bottleneck. Importantly, this pattern persists even in large transformers \textit{with skip connections} (Appendix \ref{svhn_moe_vit}).

\subsection{Expert Specialization and Confidence}
\label{Mechanistic}

We next analyze internal routing dynamics to identify the causal drivers of \textsc{Excitation}'s gains, specifically how modulation influences expert specialization and routing stability.

\begin{figure}[ht]
     \centering
     \begin{subfigure}[b]{0.49\linewidth}
         \centering
         \includegraphics[width=\linewidth]{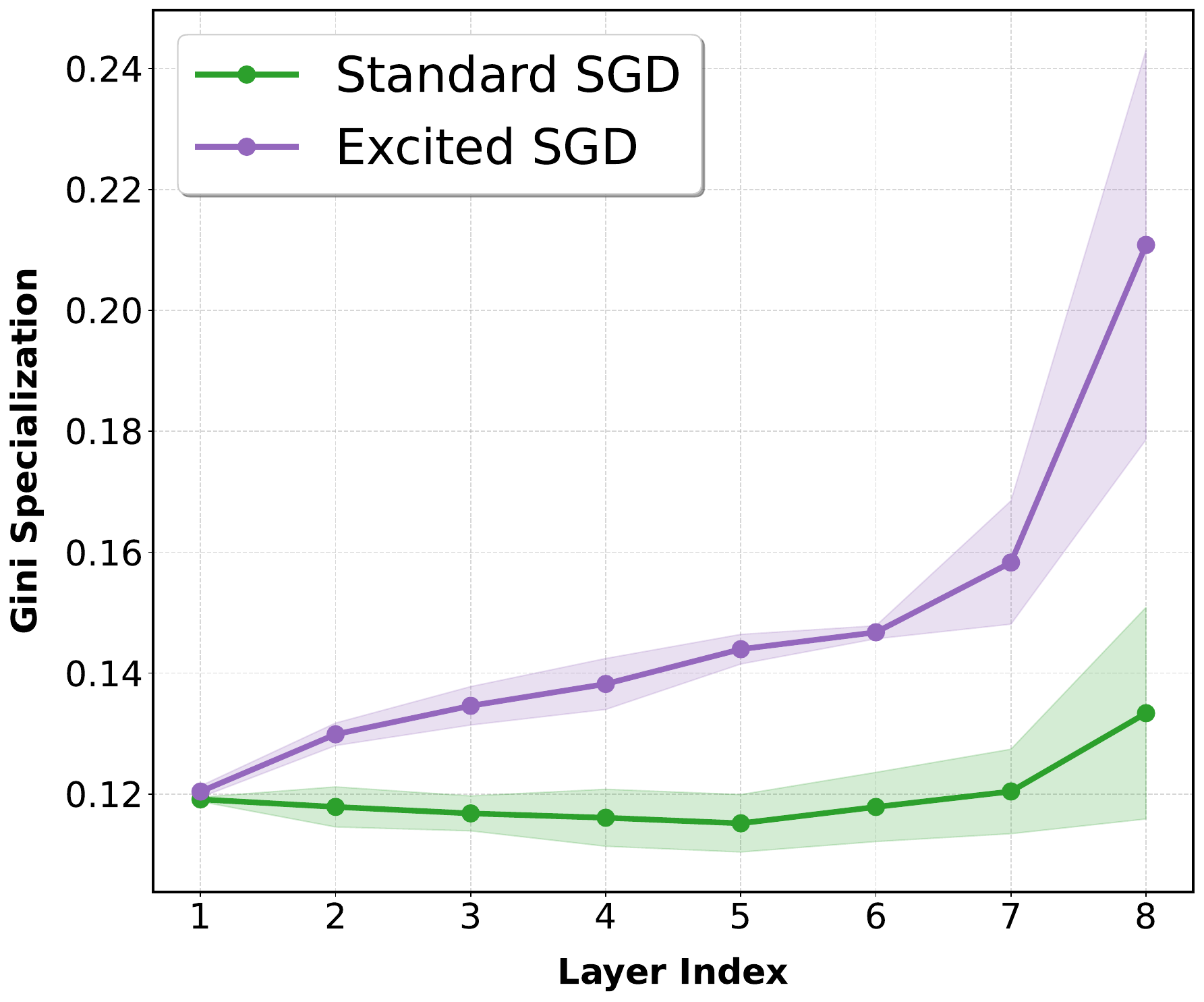}
         \caption{SGD Specialization (8 Layers)}
     \end{subfigure}
     \hfill
     \begin{subfigure}[b]{0.49\linewidth}
         \centering
         \includegraphics[width=\linewidth]{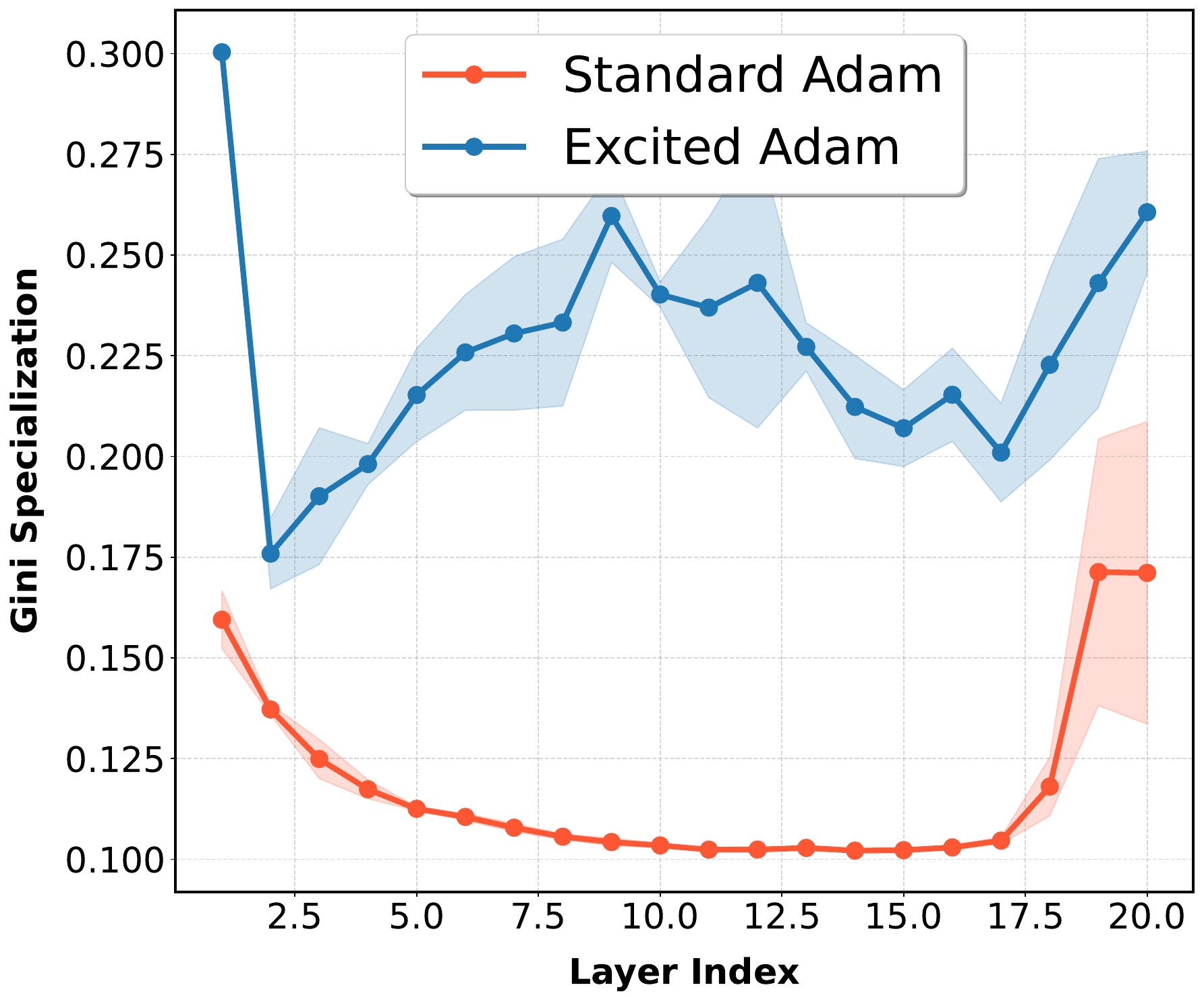}
         \caption{Adam Specialization (20 Layers)}         
     \end{subfigure}
     \caption{\textbf{Routing Specialization by Depth.} Gini coefficients of expert utilization. While standard optimizers exhibit a specialization ``dip'' in intermediate layers, \textsc{Excitation} maintains high selectivity and sharper routing across all network depths.}
     \label{fig:gini_spec}
\end{figure}

\paragraph{Specialization and Routing Dynamics} 
To quantify the impact of \textsc{Excitation} on conditional computation, we define the \textit{specialization score} $\mathcal{S}_{\ell}$ for layer $\ell$ using the Gini-Simpson Index. Let $S \in \mathbb{R}^{N \times C}$ be a class-expert utilization matrix where $S_{n,c}$ denotes the frequency of expert $n$ being selected for samples of class $c$. We define the class-conditional probability distribution for expert $n$ as $p_{n,c} = S_{n,c} / \sum_{j=1}^{C} S_{n,j}$. The layer specialization is the mean concentration across all active experts:
\begin{equation}
    \mathcal{S}_{\ell} = \frac{1}{|N_{active}|} \sum_{n \in N_{active}} \sum_{c=1}^{C} p_{n,c}^2
\end{equation}
where $\mathcal{S}_{\ell} \in [1/C, 1]$. As shown in Fig.~\ref{fig:gini_spec}, standard optimizers exhibit severe specialization collapse in intermediate layers ($\mathcal{S} \approx 0.1$ for $C=10$), indicating a trend toward feature homogenization where experts learn redundant, overlapping representations. In contrast, \textsc{Excitation} maintains high expert selectivity throughout the network. By amplifying high-consensus experts, the framework prevents the loss of routing diversity. For SGD, \textsc{Excitation} facilitates monotonic increase in specialization with depth; for Adam, it recovers the ``specialization dip'' entirely, maintaining distinct roles where standard Adam fails. These results demonstrate that active update modulation partitions sparse capacity into distinct experts rather than redundant features.

\begin{figure}[h]
     \centering
     \begin{subfigure}[b]{0.49\linewidth}
         \centering
         \includegraphics[width=\linewidth]{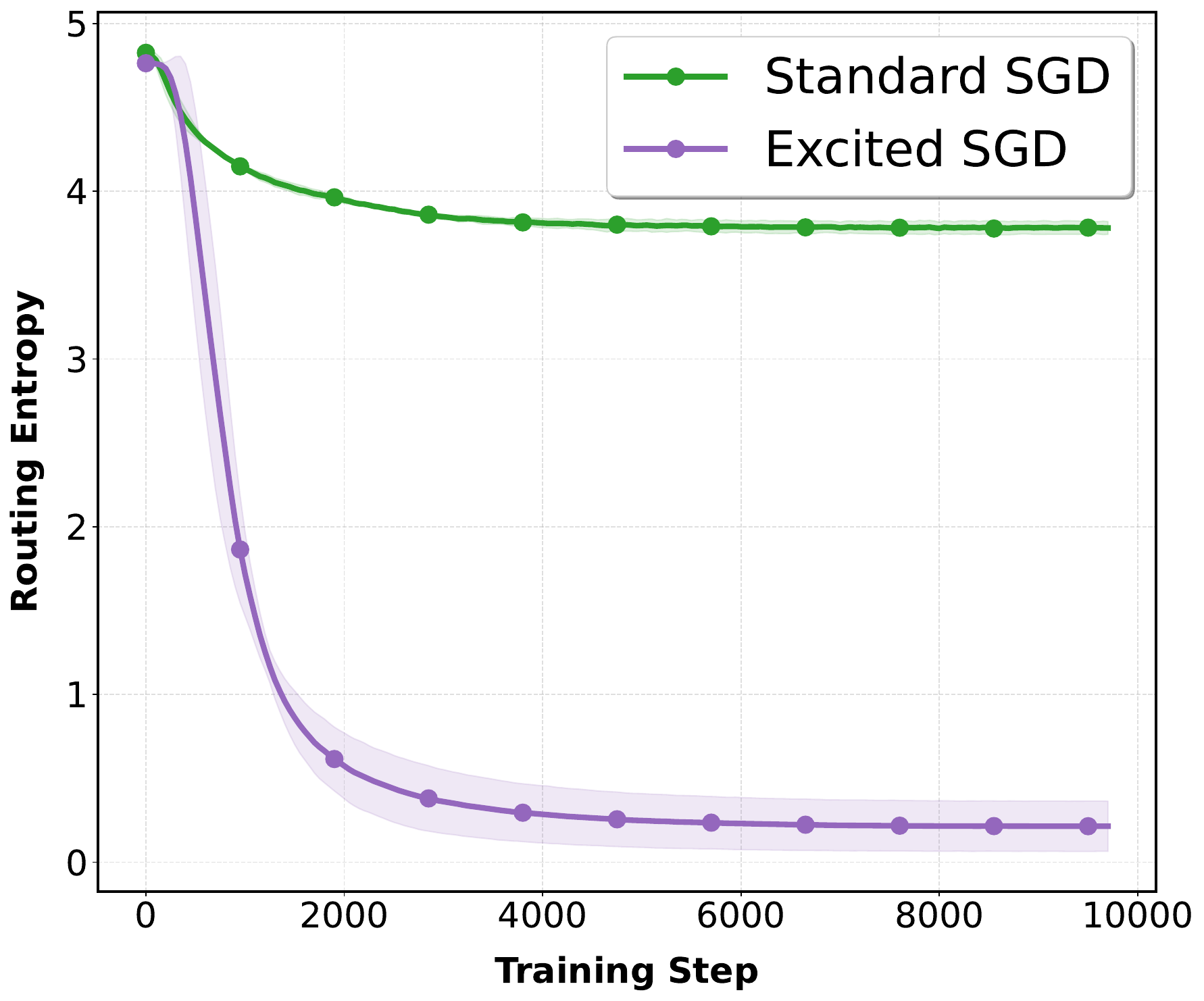}
         \caption{SGD Entropy (8 Layers)}
     \end{subfigure}
     \hfill
     \begin{subfigure}[b]{0.49\linewidth}
         \centering
         \includegraphics[width=\linewidth]{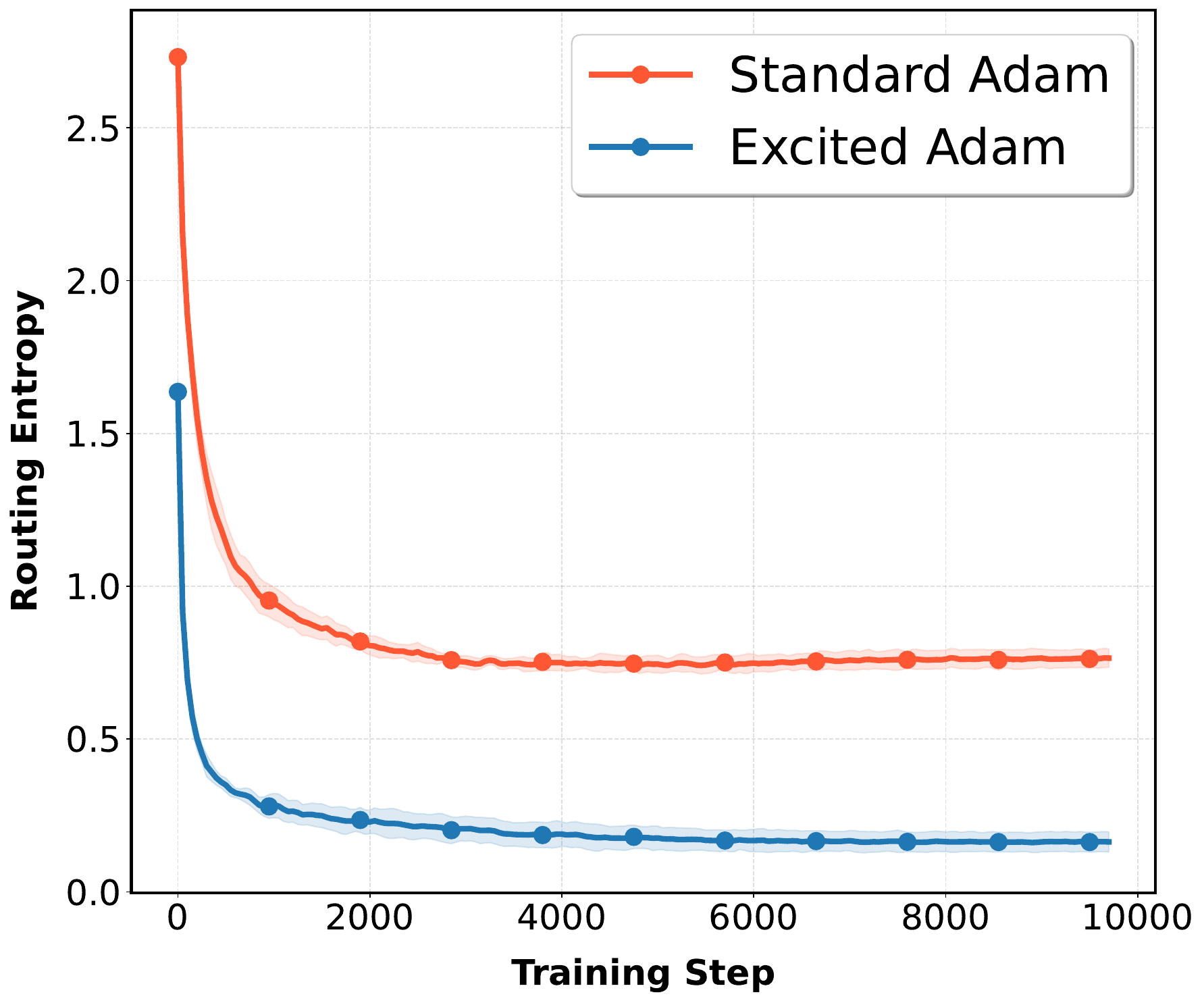}
         \caption{Adam Entropy (20 Layers)}         
     \end{subfigure}
     \caption{\textbf{Evolution of Routing Entropy.} Comparison of mean entropy across training. \textsc{Excitation} exhibits faster decay and a lower entropy floor, demonstrating that update modulation enables earlier convergence to highly specialized states.}
     \label{fig:routing_entropy}
\end{figure}

\paragraph{Routing Entropy and Convergence Dynamics}
We evaluate routing convergence via Mean Routing Entropy ($H = -\mathbb{E} [\sum_{i} p_i \log p_i]$), where lower values signify increased router confidence. As shown in Fig.~\ref{fig:routing_entropy}, \textsc{Excitation} accelerates entropy decay and achieves a lower steady-state floor compared to standard optimizers. This rapid initial collapse suggests that competitive update dynamics enable the router to identify optimal expert-task assignments much earlier than baselines. By converging to a more deterministic routing state, \textsc{Excitation} mitigates the gradient noise inherent in redundant expert utilization, directly facilitating the superior convergence trajectories observed.

\subsection{Scaling to Transformer-based MoEs}
We evaluate \textsc{Excitation} on Vision Transformers (ViT) MoEs and autoregressive GPT MoEs, studying the trade-off between expert granularity and capacity. In these models, experts correspond to standard transformer MLP blocks, and training uses the standard load-balancing loss.

\paragraph{Vision MoEs}

\begin{figure}[ht]
     \centering
     \begin{subfigure}[b]{0.48\linewidth}
          \centering
          \includegraphics[width=\linewidth]{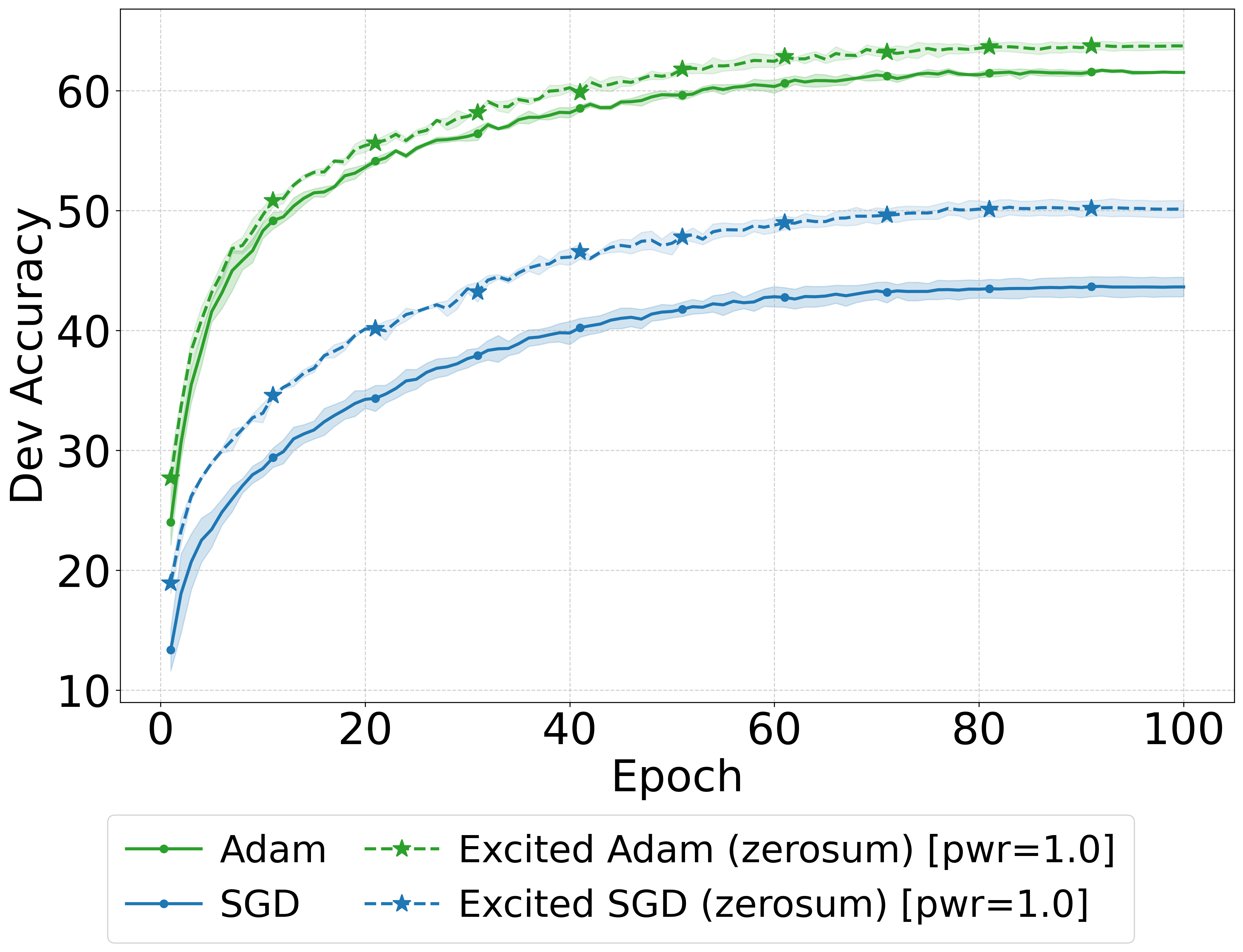}
          \caption{High Granularity (64 Experts, $L=4$, $d=128$)}
          \label{fig:vit_large_experts}
     \end{subfigure}
     \hfill
     \begin{subfigure}[b]{0.48\linewidth}
          \centering
          \includegraphics[width=\linewidth]{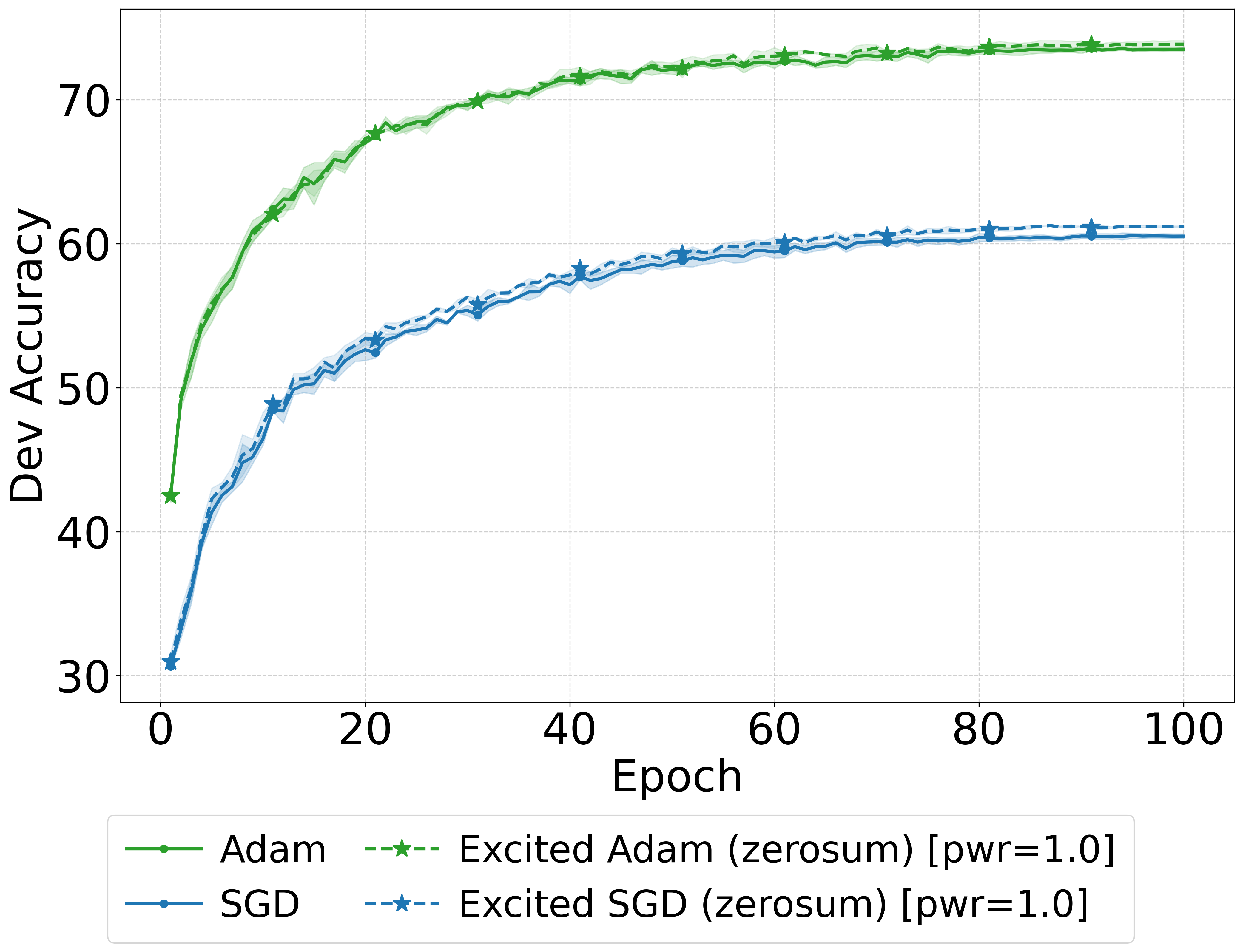}
          \caption{High Capacity (32 Experts, $L=2$, $d=512$)}
          \label{fig:vit_small_experts}
     \end{subfigure}
     \caption{\textbf{Granularity vs. Capacity (MoE-ViT).} (a) \textbf{Structural Rescue:} In ``tall and skinny'' regimes \textsc{Excitation} mitigates gradient dilution, enabling accuracy leap over vanilla plateaus. (b) \textbf{Safety Net:} In high-capacity regimes \textsc{Excitation} matches tuned baselines, ensuring robustness without over-regularization.}
     \label{fig:moe_vit_scaling}
\end{figure}

Experiments across CIFAR-10/100 and SVHN reveal a ``structural rescue'' effect in high-granularity regimes. In a 17M parameter ViT-MoE (64 experts, $L=4$, $d=128$, $TopK=2$), extreme sparsity with small experts causes standard optimizers to plateau prematurely due to specialization collapse. \textsc{Excitation} effectively catalyzes functional differentiation (Fig.~\ref{fig:moe_vit_scaling}), improving Adam's final accuracy from 61.53\% to 63.74\% and yielding a substantial 6.5\% absolute gain for SGD (43.64\% to 50.14\%).

Conversely, in high-capacity regimes (32 Experts, $L=2$, $d=512$, $TopK=8$), where routing is more stable, \textsc{Excitation} only marginally exceeds the performance of tuned Adam and SGD baselines (e.g., Adam: 73.50\% $\to$ 73.85\%). This confirms its behavior as a passive ``safety net'' that avoids over-regularization when a healthy equilibrium is already reached. Detailed results for CIFAR-100 and SVHN are provided in Appendix~\ref{moe_vit_extended}. Specifically, on SVHN, we observe a secondary instance of structural rescue: while standard SGD remains trapped in a sub-optimal plateau at approximately 20\% accuracy, \textsc{Excitation} enables the model to break out of this state, reaching $\sim$65\%. This highlights the framework's ability to navigate the complex, non-convex landscape of sparse routing where traditional first-order methods fail to establish functional signal paths.

\paragraph{Language MoEs}


\begin{table}[t]
\centering
\caption{\textbf{Relative Perplexity ($\Delta$PPL) Improvement across 60 GPT-MoE Architectures.} Values denote absolute PPL reduction ($\text{PPL}_{\text{Adam}} - \text{PPL}_{\text{Excited}}$); positive values indicate improved (lower) perplexity over the baseline.}
\label{tab:results_final}
\small
\resizebox{\columnwidth}{!}{%
\begin{tabular}{llccccc}
\toprule
\textbf{N} & \textbf{k} & \textbf{$d_{ff}$=256} & \textbf{$d_{ff}$=384} & \textbf{$d_{ff}$=512} & \textbf{$d_{ff}$=768} & \textbf{$d_{ff}$=1024} \\
\midrule
\textbf{4} & 2 & \color{teal}{77M $\mid$ +0.64} & \color{teal}{83M $\mid$ +0.42} & \color{teal}{89M $\mid$ +0.15} & \color{teal}{102M $\mid$ +0.75} & \color{teal}{115M $\mid$ +0.55} \\
\midrule
\textbf{8} & 2 & \color{teal}{89M $\mid$ +0.72} & \color{teal}{102M $\mid$ +0.83} & \color{teal}{115M $\mid$ +0.64} & \color{teal}{140M $\mid$ +0.92} & \color{teal}{165M $\mid$ +1.15} \\
 & 4 & \color{teal}{89M $\mid$ +0.48} & \color{teal}{102M $\mid$ +0.31} & \color{teal}{115M $\mid$ +0.19} & \color{teal}{140M $\mid$ +0.29} & \color{teal}{165M $\mid$ +0.47} \\
\midrule
\textbf{16} & 2 & \color{teal}{115M $\mid$ +1.18} & \color{teal}{140M $\mid$ +0.49} & \color{teal}{165M $\mid$ +0.27} & \color{teal}{216M $\mid$ +0.35} & \color{teal}{266M $\mid$ +0.67} \\
 & 4 & \color{teal}{115M $\mid$ +0.75} & \color{teal}{140M $\mid$ +0.70} & \color{teal}{165M $\mid$ +0.51} & \color{teal}{216M $\mid$ +0.61} & \color{teal}{266M $\mid$ +0.65} \\
 & 8 & \color{teal}{115M $\mid$ +0.45} & \color{teal}{140M $\mid$ +0.35} & \color{teal}{165M $\mid$ +0.15} & \color{teal}{216M $\mid$ +0.39} & \color{teal}{266M $\mid$ +0.39} \\
\midrule
\textbf{32} & 2 & \color{teal}{165M $\mid$ +0.27} & \color{teal}{216M $\mid$ +0.09} & \color{teal}{266M $\mid$ +0.27} & \color{teal}{367M $\mid$ +0.10} & \color{teal}{468M $\mid$ +0.44} \\
 & 4 & \color{teal}{165M $\mid$ +0.77} & \color{teal}{216M $\mid$ +0.18} & \color{teal}{266M $\mid$ +0.38} & \color{teal}{367M $\mid$ +0.55} & \color{teal}{468M $\mid$ +0.65} \\
 & 8 & \color{teal}{165M $\mid$ +0.06} & \color{teal}{216M $\mid$ +0.21} & \color{teal}{266M $\mid$ +0.49} & \color{teal}{367M $\mid$ +0.27} & \color{teal}{468M $\mid$ +0.28} \\
\midrule
\textbf{64} & 2 & \color{teal}{266M $\mid$ +0.48} & \color{teal}{367M $\mid$ +0.72} & \color{teal}{468M $\mid$ +0.28} & \color{teal}{670M $\mid$ +0.46} & \color{teal}{871M $\mid$ +0.47} \\
 & 4 & \color{teal}{266M $\mid$ +1.12} & \color{teal}{367M $\mid$ +0.75} & \color{teal}{468M $\mid$ +0.69} & \color{teal}{670M $\mid$ +0.72} & \color{teal}{871M $\mid$ +1.05} \\
 & 8 & \color{teal}{266M $\mid$ +0.50} & \color{teal}{367M $\mid$ +0.49} & \color{teal}{468M $\mid$ +0.50} & \color{teal}{670M $\mid$ +0.19} & \color{teal}{871M $\mid$ +0.45} \\
\bottomrule
\end{tabular}%
}
\end{table}

\begin{figure}[ht]
    \centering
    \includegraphics[width=\linewidth]{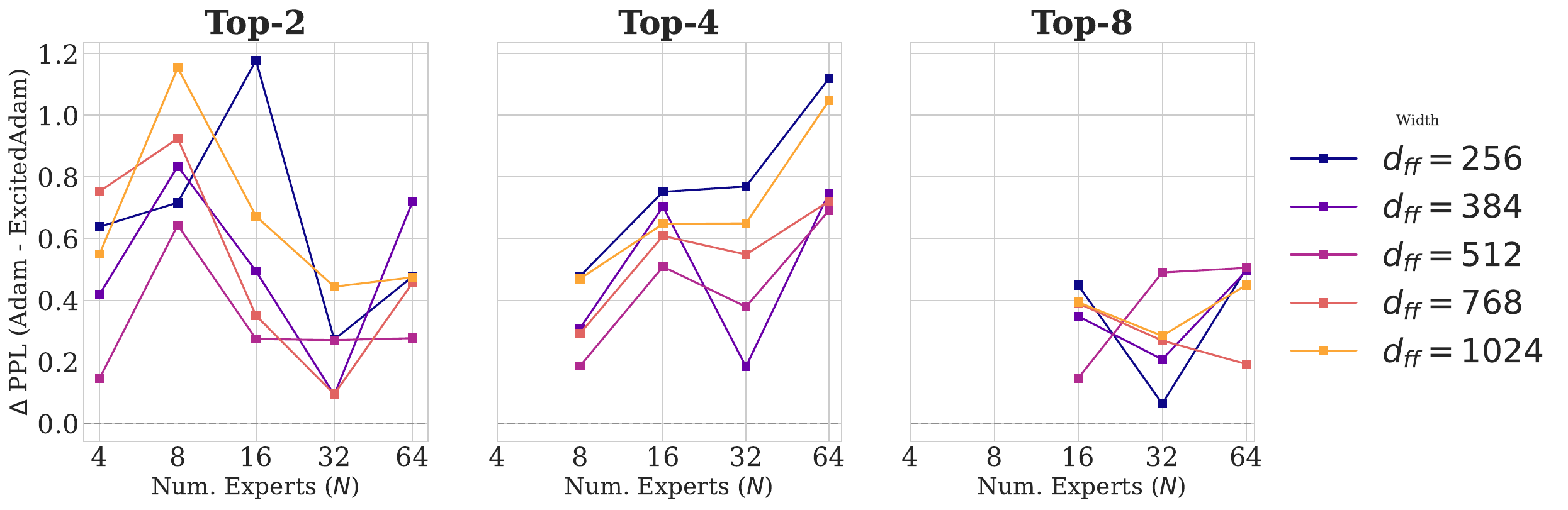} 
    \caption{\textbf{GPT-MoE Scaling Dynamics.} Relative improvement ($\Delta$PPL) over Adam as a function of the number of experts ($N$) for different routing densities ($k$) and hidden widths ($d_{ff}$).}
    \label{fig:moe_gpt_scaling}
\end{figure}

To assess the generalizability of our framework to generative tasks, we conduct an extensive sweep of 60 GPT-MoE architectures on WikiText-103, with model sizes ranging from 77M to 871M parameters. Each variant is trained for 5 epochs ($\approx$517M tokens); while compute-constrained relative to large-scale pre-training, this provides a sufficient window to observe stable convergence trends and a clear separation between optimizer variants.

As shown in Table~\ref{tab:results_final} and Fig \ref{fig:moe_gpt_scaling}, \textsc{Excitation} yields consistent perplexity (PPL) gains across the spectrum. The data reveals a complex, non-linear relationship between expert count ($N$) and Top-$k$ routing density:

\begin{itemize}[leftmargin=*, nosep]

    \item \textbf{\textsc{Excitation} utility peaks at specific structural ``stress points'' rather than scaling linearly.} The $\Delta$PPL improvement exhibits non-monotonic behavior, with a sharp initial peak at $N=8$ for Top-2 across almost all widths ($d_{ff}$). This suggests that the framework provides the most aggressive reinforcement at the exact transition where the baseline Adam first struggles to manage expert branching.
    
    \item \textbf{High-capacity models ($N=64$) exhibit a consistent ``rebound'' in gains.} Across all routing densities ($k \in \{2, 4, 8\}$), we observe a significant upward trend in $\Delta$PPL when moving from $N=32$ to $N=64$. As the routing search space becomes massive, \textsc{Excitation} becomes essential to prevent signal dilution, reaching peaks of $\approx 1.1$ $\Delta$PPL at the largest scale.

    \item \textbf{Small-expert architectures ($d_{ff}=256$) are uniquely resilient and benefit longer from reinforcement.} Unlike larger-expert configurations that regress at $N=16$ (Top-2), the $d_{ff}=256$ variant continues to peak, suggesting that smaller experts require sustained update modulation to maintain routing specialization.

    \item \textbf{\textsc{Excitation} substitutes for width-driven stability in deep routing.} In the Top-4, $N=32$ regime, larger experts ($d_{ff}=1024$) naturally resist performance dips; however, \textsc{Excitation} enables the smallest experts ($d_{ff}=256$) to achieve comparable gains ($\approx 1.1$), effectively closing the stability gap between disparate widths.

    \item \textbf{Gains are inversely correlated with routing density ($k$).} The highest relative improvements occur in sparser regimes ($k=2$ and $k=4$). In high-consensus settings ($k=8$), gains are more modest ($< 0.6$), confirming that the framework's primary value lies in enforcing differentiation when the routing signal is most sparse.
    
    \item \textbf{\textsc{Excitation} provides a universal performance floor} Across all 60 configurations (77M--871M parameters), \textsc{Excitation} \textit{never underperformed the baseline}. This establishes it as a robust enhancement for generative MoEs, regardless of expert size, count, or sparsity.
\end{itemize}

These results demonstrate that \textsc{Excitation} effectively stabilizes the capacity-routing trade-off across scales.

\subsection{Sensitivity Analysis}
\label{sensitivity_sec}
We evaluate \textsc{Excitation}'s operational boundaries and interaction with standard hyperparameters on the TopK MLP.

\paragraph{Learning Rate Stability}
We evaluate \textsc{Excitation} sensitivity across a sweep from $\eta = 10^{-4}$ up to divergence (Table~\ref{tab:combined-lr-analysis}). While vanilla Adam and SGD attain narrow point-estimate peaks at $\eta=0.002$ and $\eta=0.3$, these optima represent highly sensitive operating regions where performance degrades rapidly upon deviation.

In contrast, \textsc{Excitation} exhibits a significantly more robust profile across the sweep, maintaining higher \textbf{mean accuracy} and reduced sensitivity to $\eta$. This is most pronounced at low-to-moderate learning rates—common in large-scale training—where \textsc{Excitation} acts as a \textbf{specialization safety net}, reinforcing expert differentiation when gradient signals are otherwise insufficient for stable structure. As $\eta$ approaches the divergence threshold, gains diminish but \textsc{Excitation} remains non-destructive, indicating it complements rather than replaces standard hyperparameter optimization.


The practical value of this robustness is underscored in our MoE ViT experiments. When using \textbf{published, community-standard hyperparameters} where exhaustive sweeps are infeasible, \textsc{Excitation} yields consistent improvements despite the baseline being highly tuned. These results demonstrate that \textsc{Excitation} \textit{reduces the tuning burden by widening the usable learning-rate window while remaining compatible with carefully optimized configurations}.

\begin{table}[ht]
\centering
\caption{\textbf{Learning Rate Sensitivity (CIFAR-10).} Comparison of Vanilla vs. Excited variants (\textit{zerosum}, \textit{positivesum}) for Adam and SGD. Bold denotes peak accuracy per learning rate.}
\label{tab:combined-lr-analysis}
\small
\begin{tabular}{l S S S}
\toprule
& \multicolumn{3}{c}{\textbf{Dev Accuracy (\%)}} \\
\cmidrule(lr){2-4}
\textbf{LR ($\eta$)} & \textbf{Vanilla} & \textbf{Excited (zs)} & \textbf{Excited (ps)} \\
\midrule
\multicolumn{4}{c}{\textit{Adam Variants}} \\
\midrule
1.00e-4 & 48.47 \pm 0.23 & 49.81 \pm 0.29 & \multicolumn{1}{c}{\textbf{50.55 $\pm$ 0.24}} \\
1.50e-4 & 49.85 \pm 0.20 & 51.22 \pm 0.15 & \multicolumn{1}{c}{\textbf{51.65 $\pm$ 0.33}} \\
2.00e-4 & 51.09 \pm 0.11 & 51.66 \pm 0.29 & \multicolumn{1}{c}{\textbf{52.63 $\pm$ 0.29}} \\
5.00e-4 & 53.69 \pm 0.63 & 52.95 \pm 0.37 & \multicolumn{1}{c}{\textbf{54.07 $\pm$ 0.41}} \\
0.001   & 55.01 \pm 0.36 & 54.26 \pm 0.10 & \multicolumn{1}{c}{\textbf{55.11 $\pm$ 0.16}} \\
0.002   & \multicolumn{1}{c}{\textbf{56.12 $\pm$ 0.22}} & 54.33 \pm 0.42 & 55.58 \pm 0.44 \\
0.003   & \multicolumn{1}{c}{\textbf{56.04 $\pm$ 0.18}} & 54.99 \pm 0.02 & 55.81 \pm 0.35 \\
\midrule
\multicolumn{4}{c}{\textit{SGD Variants}} \\
\midrule
0.01  & 41.01 \pm 0.33 & 46.58 \pm 0.32 & \multicolumn{1}{c}{\textbf{46.59 $\pm$ 0.68}} \\
0.015 & 43.50 \pm 0.13 & \multicolumn{1}{c}{\textbf{48.03 $\pm$ 0.07}} & 47.99 \pm 0.22 \\
0.02  & 45.10 \pm 0.22 & \multicolumn{1}{c}{\textbf{49.01 $\pm$ 0.14}} & 48.94 \pm 0.19 \\
0.05  & 48.55 \pm 0.46 & 50.49 \pm 0.32 & \multicolumn{1}{c}{\textbf{50.90 $\pm$ 0.10}} \\
0.1   & 50.91 \pm 0.22 & 51.47 \pm 0.38 & \multicolumn{1}{c}{\textbf{51.61 $\pm$ 0.41}} \\
0.2   & 52.43 \pm 0.20 & 52.44 \pm 0.05 & \multicolumn{1}{c}{\textbf{52.65 $\pm$ 0.30}} \\
0.3   & \multicolumn{1}{c}{\textbf{53.67 $\pm$ 0.08}} & 52.70 \pm 0.16 & 53.02 \pm 0.23 \\
\bottomrule
\end{tabular}
\end{table}

\paragraph{Sparsity and Batch Size}
\begin{table}[ht]
\centering
\caption{\textbf{Sparsity Analysis (CIFAR-10).} Comparison between Vanilla and Excited (\textit{zerosum}) variants. $\Delta$ denotes the absolute percentage improvement over the baseline.}
\label{tab:sparsity-analysis-combined}
\small
\begin{tabular}{c S S S}
\toprule
\textbf{Sparsity} & \textbf{Vanilla (\%)} & \textbf{Excited (\%)} & \textbf{$\Delta$} \\
\midrule
\multicolumn{4}{c}{\textit{Adam Variants}} \\
\midrule
90\% & 48.47 \pm 0.23 & \multicolumn{1}{c}{\textbf{49.81 $\pm$ 0.29}} & \color{teal}{+1.34} \\
70\% & 54.01 \pm 0.15 & \multicolumn{1}{c}{\textbf{54.81 $\pm$ 0.28}} & \color{teal}{+0.80} \\
50\% & \multicolumn{1}{c}{\textbf{55.49 $\pm$ 0.31}} & 55.34 \pm 0.08 & \color{red}{-0.15} \\
30\% & 55.79 \pm 0.18 & \multicolumn{1}{c}{\textbf{55.86 $\pm$ 0.45}} & \color{teal}{+0.07} \\
10\% & \multicolumn{1}{c}{\textbf{56.08 $\pm$ 0.40}} & 56.00 \pm 0.15 & \color{red}{-0.08} \\
\midrule
\multicolumn{4}{c}{\textit{SGD Variants}} \\
\midrule
90\% & 41.01 \pm 0.33 & \multicolumn{1}{c}{\textbf{46.58 $\pm$ 0.32}} & \color{teal}{+5.57} \\
70\% & 48.61 \pm 0.56 & \multicolumn{1}{c}{\textbf{50.17 $\pm$ 0.42}} & \color{teal}{+1.56} \\
50\% & 50.66 \pm 0.39 & \multicolumn{1}{c}{\textbf{51.55 $\pm$ 0.21}} & \color{teal}{+0.89} \\
30\% & 51.32 \pm 0.47 & \multicolumn{1}{c}{\textbf{51.40 $\pm$ 0.51}} & \color{teal}{+0.08} \\
10\% & 51.47 \pm 0.36 & \multicolumn{1}{c}{\textbf{51.51 $\pm$ 0.40}} & \color{teal}{+0.04} \\
\bottomrule
\end{tabular}
\end{table}
\begin{table}[ht]
\centering
\caption{\textbf{Batch Size Sensitivity (CIFAR-10).} Comparison between Vanilla and Excited (\textit{zerosum}) variants. $\Delta$ denotes absolute percentage improvement.
}
\label{tab:batch-size-analysis}
\small
\begin{tabular}{c S S S}
\toprule
\textbf{Batch Size} & \textbf{Vanilla (\%)} & \textbf{Excited (\%)} & \textbf{$\Delta$} \\
\midrule
\multicolumn{4}{c}{\textit{Adam Variants}} \\
\midrule
16  & \multicolumn{1}{c}{\textbf{53.54 $\pm$ 0.26}} & 53.41 \pm 0.16 & \color{red}{-0.13} \\
32  & 52.83 \pm 0.11 & \multicolumn{1}{c}{\textbf{52.87 $\pm$ 0.23}} & \color{teal}{+0.04} \\
64  & 51.92 \pm 0.33 & \multicolumn{1}{c}{\textbf{52.08 $\pm$ 0.34}} & \color{teal}{+0.16} \\
128 & 50.77 \pm 0.07 & \multicolumn{1}{c}{\textbf{51.69 $\pm$ 0.23}} & \color{teal}{+0.92} \\
256 & 49.42 \pm 0.44 & \multicolumn{1}{c}{\textbf{50.55 $\pm$ 0.24}} & \color{teal}{+1.13} \\
512 & 48.47 \pm 0.23 & \multicolumn{1}{c}{\textbf{49.81 $\pm$ 0.29}} & \color{teal}{+1.34} \\
\midrule
\multicolumn{4}{c}{\textit{SGD Variants}} \\
\midrule
16  & \multicolumn{1}{c}{\textbf{53.73 $\pm$ 0.25}} & 53.21 \pm 0.46 & \color{red}{-0.52} \\
32  & 51.73 \pm 0.22 & \multicolumn{1}{c}{\textbf{52.52 $\pm$ 0.06}} & \color{teal}{+0.79} \\
64  & 50.52 \pm 0.22 & \multicolumn{1}{c}{\textbf{51.54 $\pm$ 0.10}} & \color{teal}{+1.02} \\
128 & 48.05 \pm 0.36 & \multicolumn{1}{c}{\textbf{50.16 $\pm$ 0.46}} & \color{teal}{+2.11} \\
256 & 45.19 \pm 0.05 & \multicolumn{1}{c}{\textbf{48.69 $\pm$ 0.16}} & \color{teal}{+3.50} \\
512 & 41.01 \pm 0.33 & \multicolumn{1}{c}{\textbf{46.58 $\pm$ 0.32}} & \color{teal}{+5.57} \\
\bottomrule
\end{tabular}
\end{table}

We observe a strong positive correlation between sparsity levels and performance gains (Table \ref{tab:sparsity-analysis-combined}). In the SGD regime, gains increase monotonically from $+0.04\%$ at 10\% sparsity to $+5.57\%$ at 90\% sparsity. This confirms that \textsc{Excitation} is uniquely suited for highly sparse environments where ``routing confusion'' is prevalent. Furthermore, effectiveness scales with batch size (Table \ref{tab:batch-size-analysis}). Smaller batches yield high-variance utilization statistics, while larger batches provide a high-fidelity ``consensus'' signal that enables more precise spatial targeting of updates. However, for memory-constrained settings requiring smaller batches, future work could explore \textit{excitation accumulation}, where expert utilization statistics are aggregated across multiple gradient accumulation steps before applying modulation, effectively decoupling the consensus window from the physical batch size.

\paragraph{Scheduler and Power ($\gamma$) Robustness}
\begin{table}[ht]
\centering
\caption{Scheduler Sensitivity Analysis on CIFAR-10. Comparison between Vanilla and Excited (zerosum) variants with and without Cosine LR scheduling. $\Delta$ represents absolute percentage improvement. 
}
\label{tab:scheduler-analysis}
\small
\begin{tabular}{l S S S}
\toprule
\textbf{Scheduler} & \textbf{Vanilla (\%)} & \textbf{Excited (\%)} & \textbf{$\Delta$} \\
\midrule
\multicolumn{4}{c}{\textit{Adam Variants}} \\
\midrule
None (Constant) & 50.85 \pm 0.25 & \multicolumn{1}{c}{\textbf{51.28 $\pm$ 0.57}} & \color{teal}{+0.43} \\
Cosine Decay    & 48.47 \pm 0.23 & \multicolumn{1}{c}{\textbf{49.81 $\pm$ 0.29}} & \color{teal}{+1.34} \\
\midrule
\multicolumn{4}{c}{\textit{SGD Variants}} \\
\midrule
None (Constant) & 44.94 \pm 0.10 & \multicolumn{1}{c}{\textbf{49.10 $\pm$ 0.55}} & \color{teal}{+4.16} \\
Cosine Decay    & 41.01 \pm 0.33 & \multicolumn{1}{c}{\textbf{46.58 $\pm$ 0.32}} & \color{teal}{+5.57} \\
\bottomrule
\end{tabular}
\end{table}
\begin{table}[ht]
\centering
\caption{Power ($p$) Sensitivity Analysis on CIFAR-10. Comparison of Vanilla and Excited variants (zerosum and positivesum) across different power values. $\Delta$ represents the absolute percentage improvement over the Vanilla baseline. 
}
\label{tab:power-analysis}
\small
\begin{tabular}{c l S S}
\toprule
\textbf{Power ($\gamma$)} & \textbf{Variant} & \textbf{Dev Accuracy (\%)} & \textbf{$\Delta$} \\
\midrule
\multicolumn{4}{c}{\textit{Adam Variants (Vanilla: 48.47 $\pm$ 0.23)}} \\
\midrule
\multirow{2}{*}{1.0} & Excited (zs) & 49.81 \pm 0.29 & \color{teal}{+1.34} \\
                     & \textbf{Excited (ps)} & \multicolumn{1}{c}{\textbf{50.55 $\pm$ 0.24}} & \color{teal}{\textbf{+2.08}} \\
\midrule
\multirow{2}{*}{2.0} & Excited (zs) & 49.47 \pm 0.64 & \color{teal}{+1.00} \\
                     & Excited (ps) & 50.45 \pm 0.42 & \color{teal}{+1.98} \\
\midrule
\multirow{2}{*}{3.0} & Excited (zs) & 49.36 \pm 0.50 & \color{teal}{+0.89} \\
                     & Excited (ps) & 50.46 \pm 0.22 & \color{teal}{+1.99} \\
\midrule
\multicolumn{4}{c}{\textit{SGD Variants (Vanilla: 41.01 $\pm$ 0.33)}} \\
\midrule
\multirow{2}{*}{1.0} & Excited (zs) & 46.58 \pm 0.32 & \color{teal}{+5.57} \\
                     & Excited (ps) & 46.59 \pm 0.68 & \color{teal}{+5.58} \\
\midrule
\multirow{2}{*}{2.0} & Excited (zs) & 46.19 \pm 0.07 & \color{teal}{+5.18} \\
                     & \textbf{Excited (ps)} & \multicolumn{1}{c}{\textbf{46.63 $\pm$ 0.42}} & \color{teal}{\textbf{+5.62}} \\
\midrule
\multirow{2}{*}{3.0} & Excited (zs) & 45.11 \pm 0.35 & \color{teal}{+4.10} \\
                     & Excited (ps) & 46.59 \pm 0.39 & \color{teal}{+5.58} \\
\bottomrule
\end{tabular}
\end{table}

\textsc{Excitation} is highly complementary to standard scheduling; peak gains occur with Cosine Decay (Table \ref{tab:scheduler-analysis}), suggesting that update ``sharpening'' becomes more critical as the learning rate decays. The framework is also robust to the power hyperparameter $\gamma$ (Table \ref{tab:power-analysis}). While $\Phi_{PS}$ is stable across $\gamma \in \{1, 2, 3\}$, $\Phi_{ZS}$ shows moderate sensitivity at higher values, where aggressive suppression can risk expert collapse. This overall stability confirms \textsc{Excitation} does not require exhaustive hyperparameter tuning, with $\gamma=1$ serving as the recommended default.

\section{Conclusions}
We introduced \textsc{Excitation}, an optimization framework that accelerates sparse architecture learning by modulating updates via batch-level expert utilization. By introducing a competitive update dynamic, \textsc{Excitation} sharpens routing specialization and resolves ``structural confusion'' in deep regimes where standard optimizers fail to establish functional signal paths. Our evaluation across diverse vision and language tasks shows consistent improvements in convergence speed and final performance, confirming the framework's efficacy as a ``specialization catalyst''.

\section*{Impact Statement}
This paper presents a framework designed to improve the training efficiency and convergence of sparse machine learning models. By optimizing the utilization of conditional computation, our work contributes to the development of more computationally efficient architectures, potentially reducing the energy requirements and carbon footprint associated with training large-scale models. While the broader societal consequences of advancing machine learning are diverse, we do not feel there are specific ethical concerns or negative societal impacts that must be uniquely highlighted for this optimization framework.

\bibliography{citations}
\bibliographystyle{icml2026}

\newpage
\appendix
\onecolumn
\section{Appendix}

\section{Foundational Convergence on CIFAR-100 and SVHN}\label{sgd_svhn_cifar100}
\begin{figure}[H]\centering\begin{subfigure}[b]{0.48\linewidth}\centering\includegraphics[width=\linewidth]{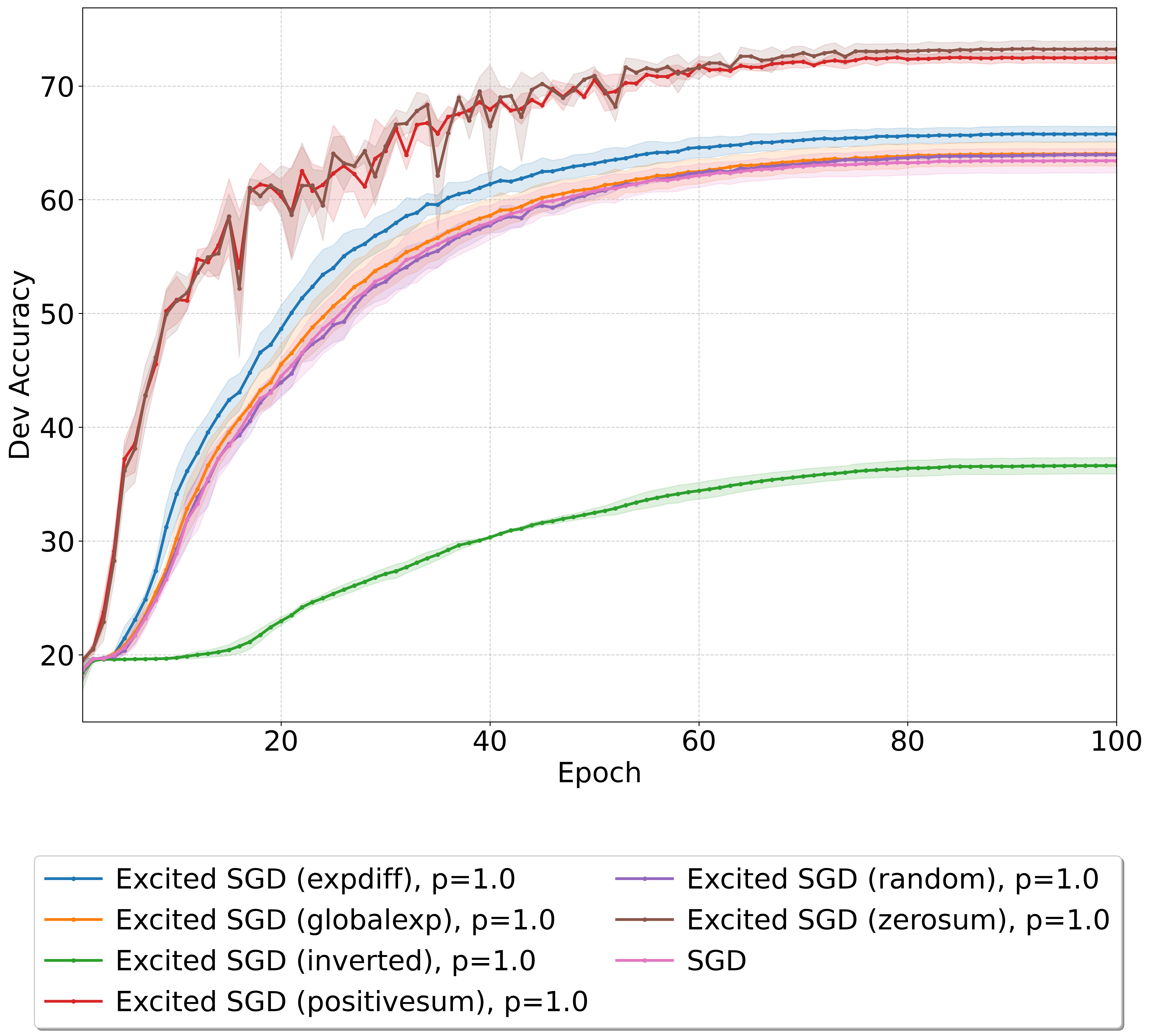}\caption{SVHN (Micro-MoE)}\label{fig:svhn_foundation_sgd}\end{subfigure}\hfill\begin{subfigure}[b]{0.48\linewidth}\centering\includegraphics[width=\linewidth]{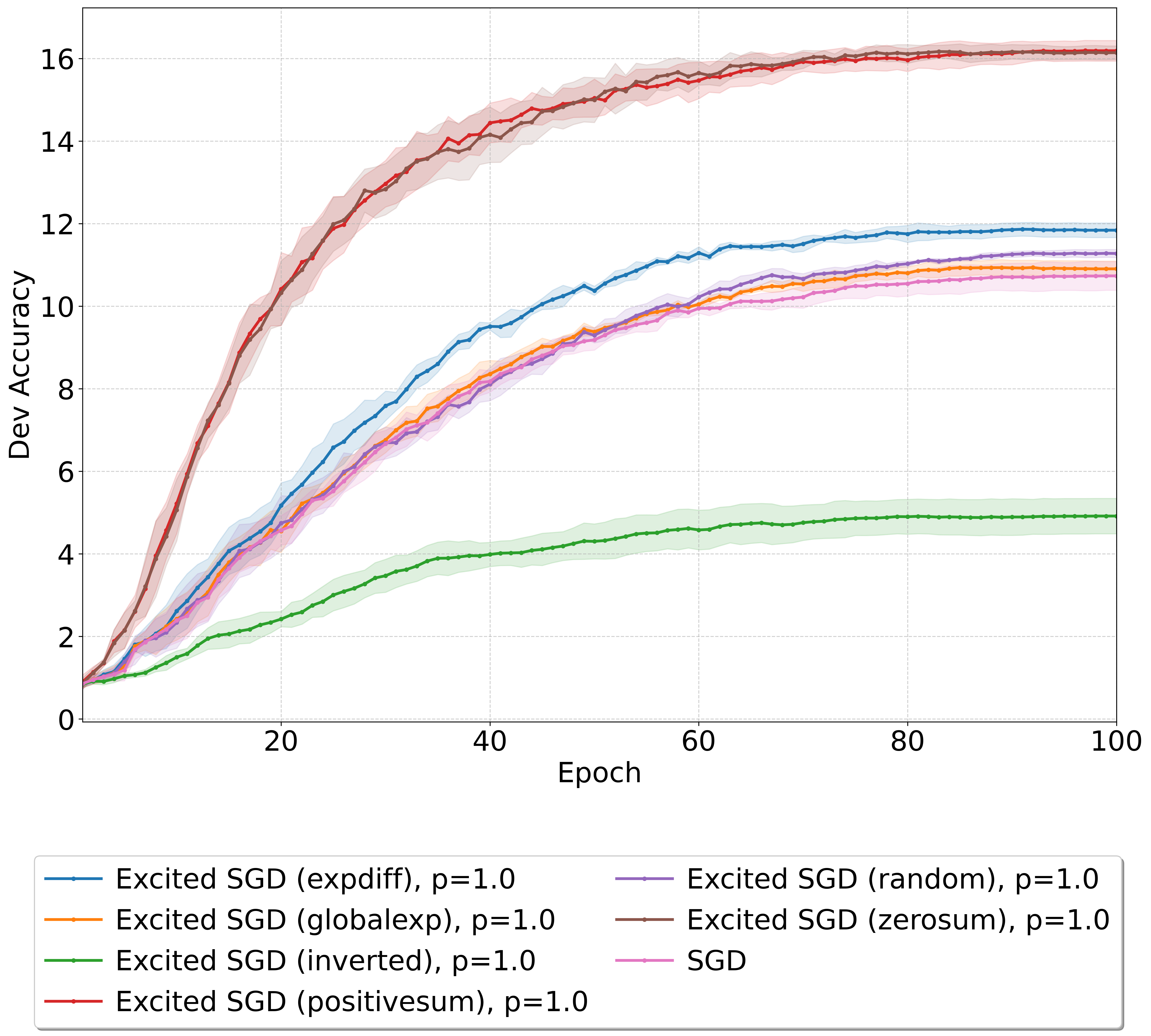}\caption{CIFAR-100 (Micro-MoE)}\label{fig:cifar100_foundation_sgd}\end{subfigure}\caption{\textbf{Foundational Benchmarks across Datasets.} Comparison of \textsc{Excitation} variants against standard SGD and controls. The consistent superiority of targeted excitation confirms that the framework's ability to resolve structural confusion is domain-agnostic.}\end{figure}

To ensure the ``Structural Rescue'' effect observed on CIFAR-10 is not an artifact of dataset simplicity, we evaluate the foundational Micro-MoE setup on CIFAR-100 and SVHN. These datasets provide a rigorous test for the framework's ability to handle increased label complexity (CIFAR-100) and varying image distributions (SVHN). As illustrated in Figures \ref{fig:svhn_foundation_sgd} and \ref{fig:cifar100_foundation_sgd}, the performance hierarchy remains remarkably consistent with our primary findings. In both cases, the targeted variants ($\Phi_{ZS}$ and $\Phi_{PS}$) significantly outperform the standard SGD baseline and the Global-Exp control. Notably, the failure of the Inverted control is even more pronounced on CIFAR-100, suggesting that as task complexity increases, the cost of ``structural confusion'' becomes higher, making the specialization signal provided by \textsc{Excitation} increasingly critical for convergence.

\section{Extended Training Dynamics Across Optimizers}
\label{different_opt_training_curves}

To supplement the terminal accuracy results provided in the main text, we visualize the complete training trajectories for a diverse suite of base optimizers equipped with the \textsc{Excitation}. These curves demonstrate that the framework's benefits are sustained throughout the optimization process, regardless of the underlying update rule.

\begin{figure}[H]
     \centering
     \begin{subfigure}[b]{0.48\linewidth}
          \centering
          \includegraphics[width=\linewidth]{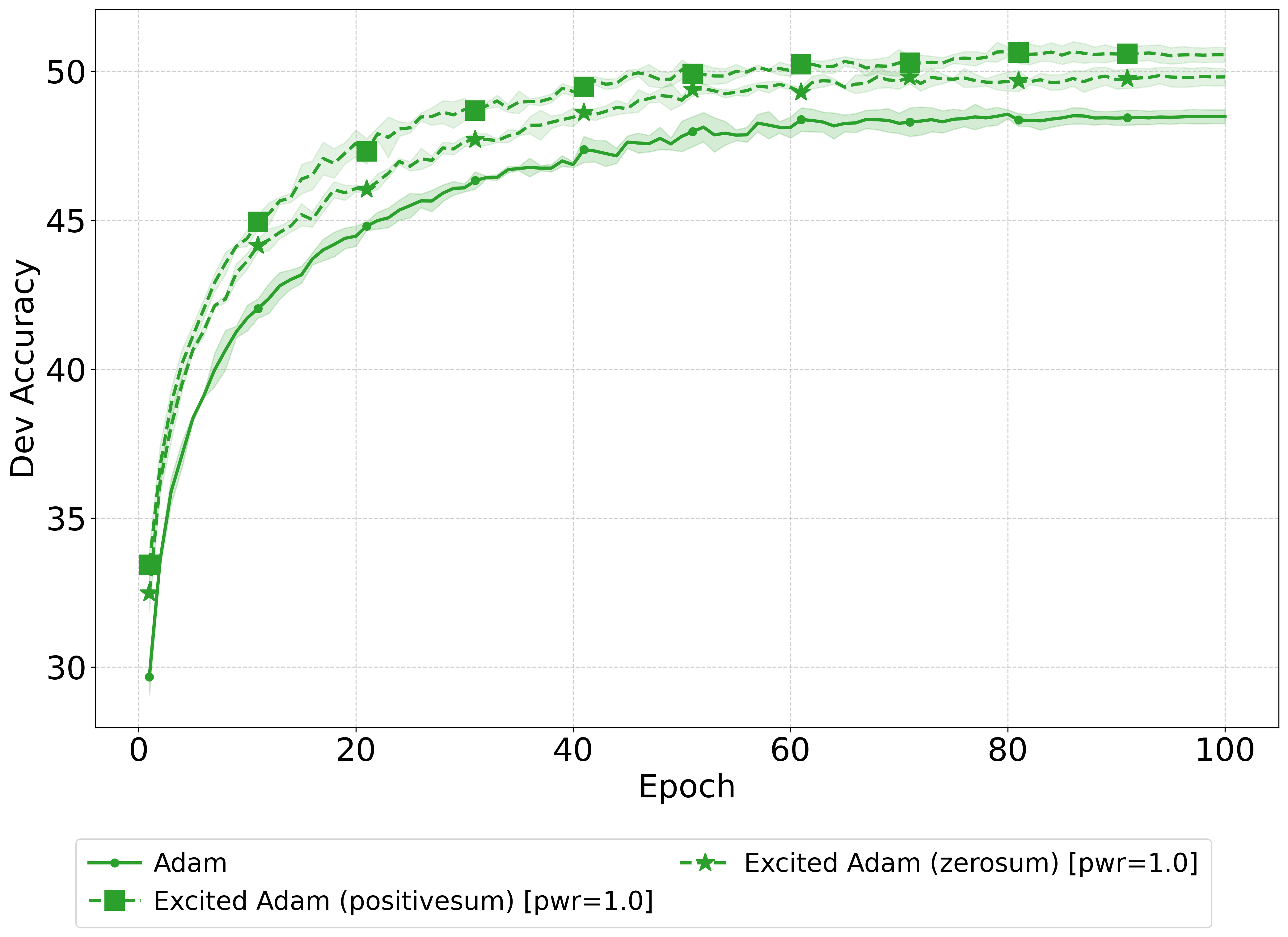}
          \caption{Adam}
          \label{fig:cifar10_adam}
     \end{subfigure}
     \hfill
     \begin{subfigure}[b]{0.48\linewidth}
          \centering
          \includegraphics[width=\linewidth]{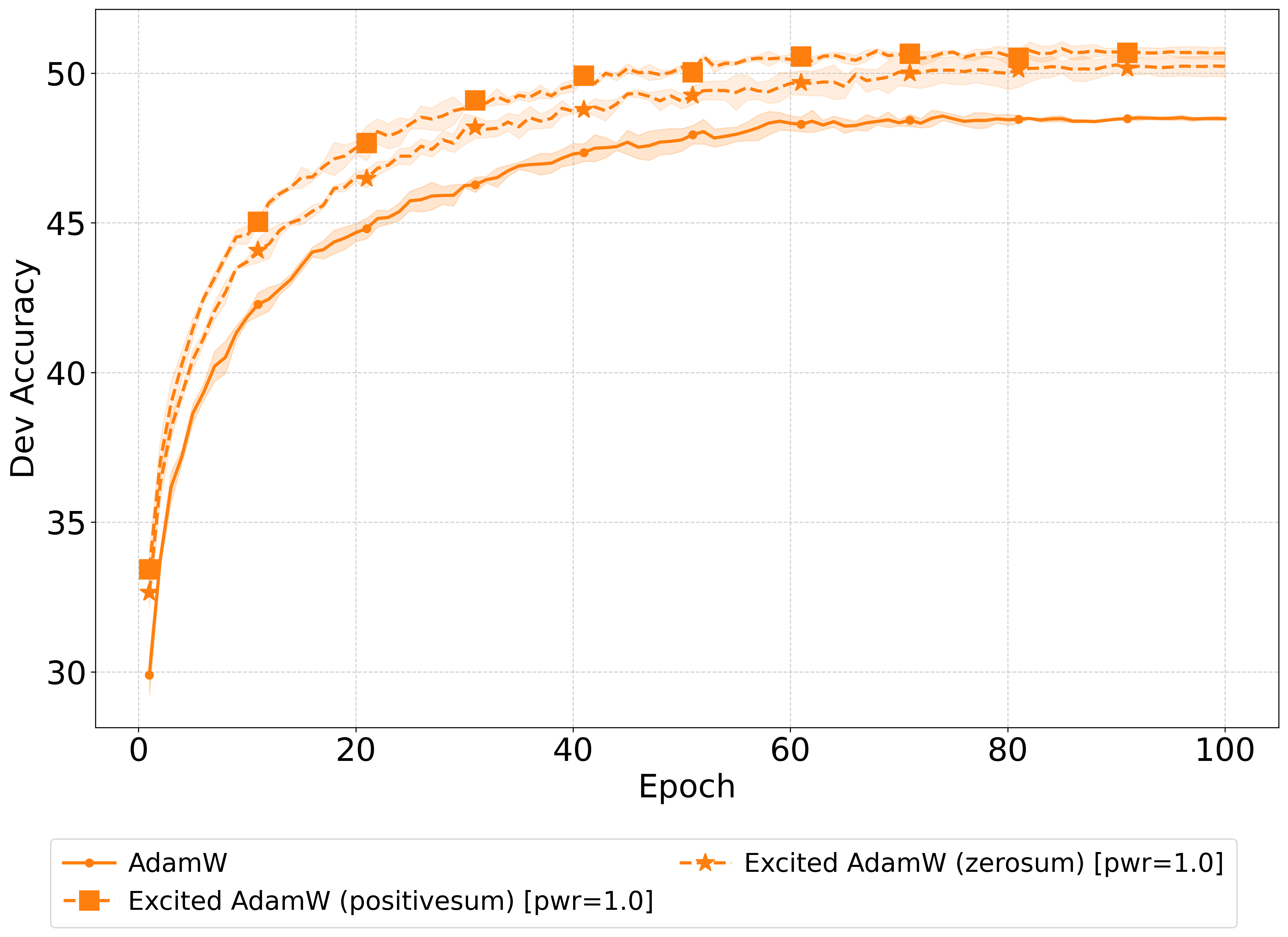}
          \caption{AdamW}
          \label{fig:cifar10_adamw}
     \end{subfigure}

     \vspace{1em}

     \begin{subfigure}[b]{0.48\linewidth}
          \centering
          \includegraphics[width=\linewidth]{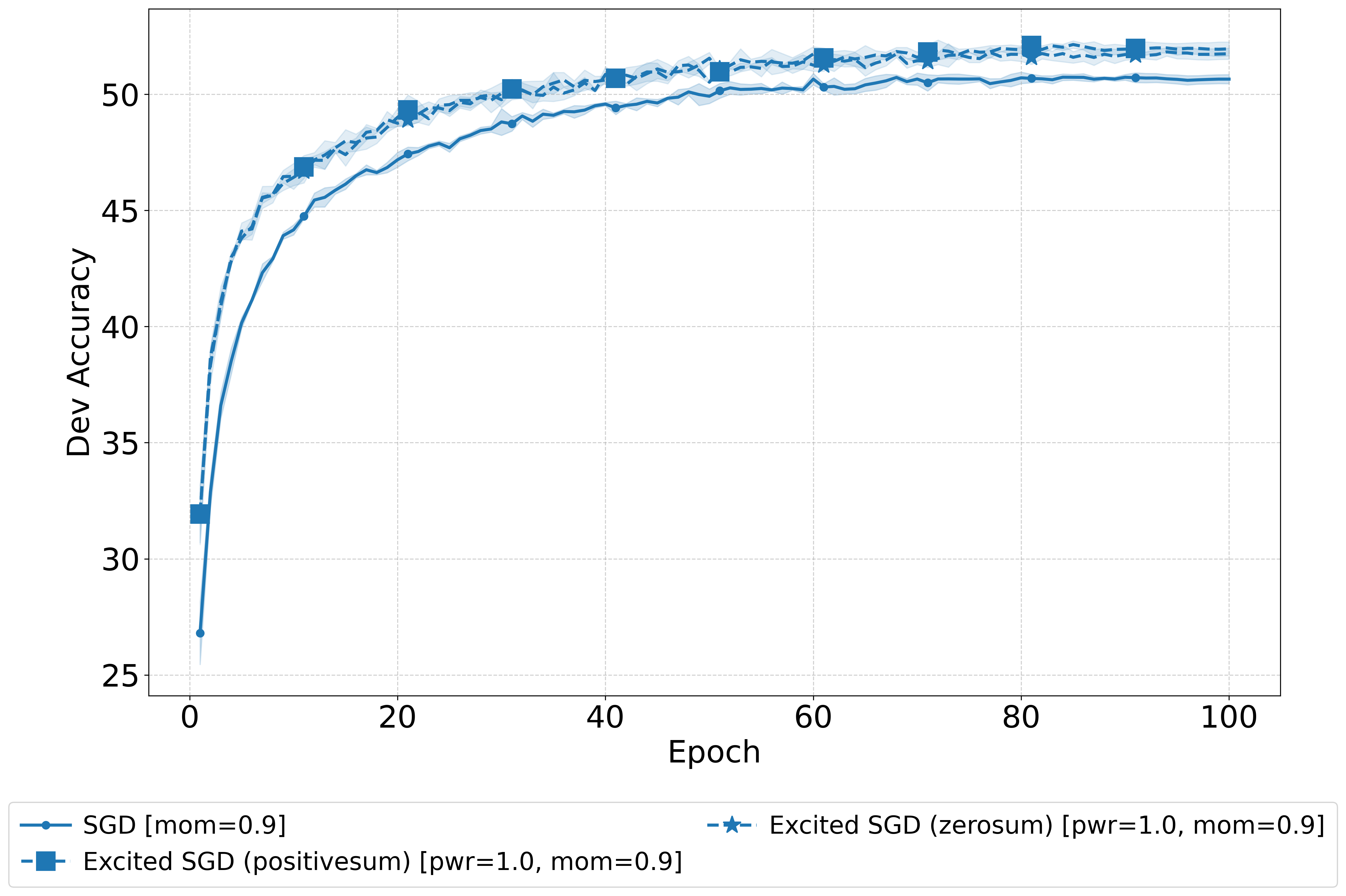}
          \caption{SGD (Momentum=0.9)}
          \label{fig:cifar10_sgd_mom}
     \end{subfigure}
     \hfill
     \begin{subfigure}[b]{0.48\linewidth}
          \centering
          \includegraphics[width=\linewidth]{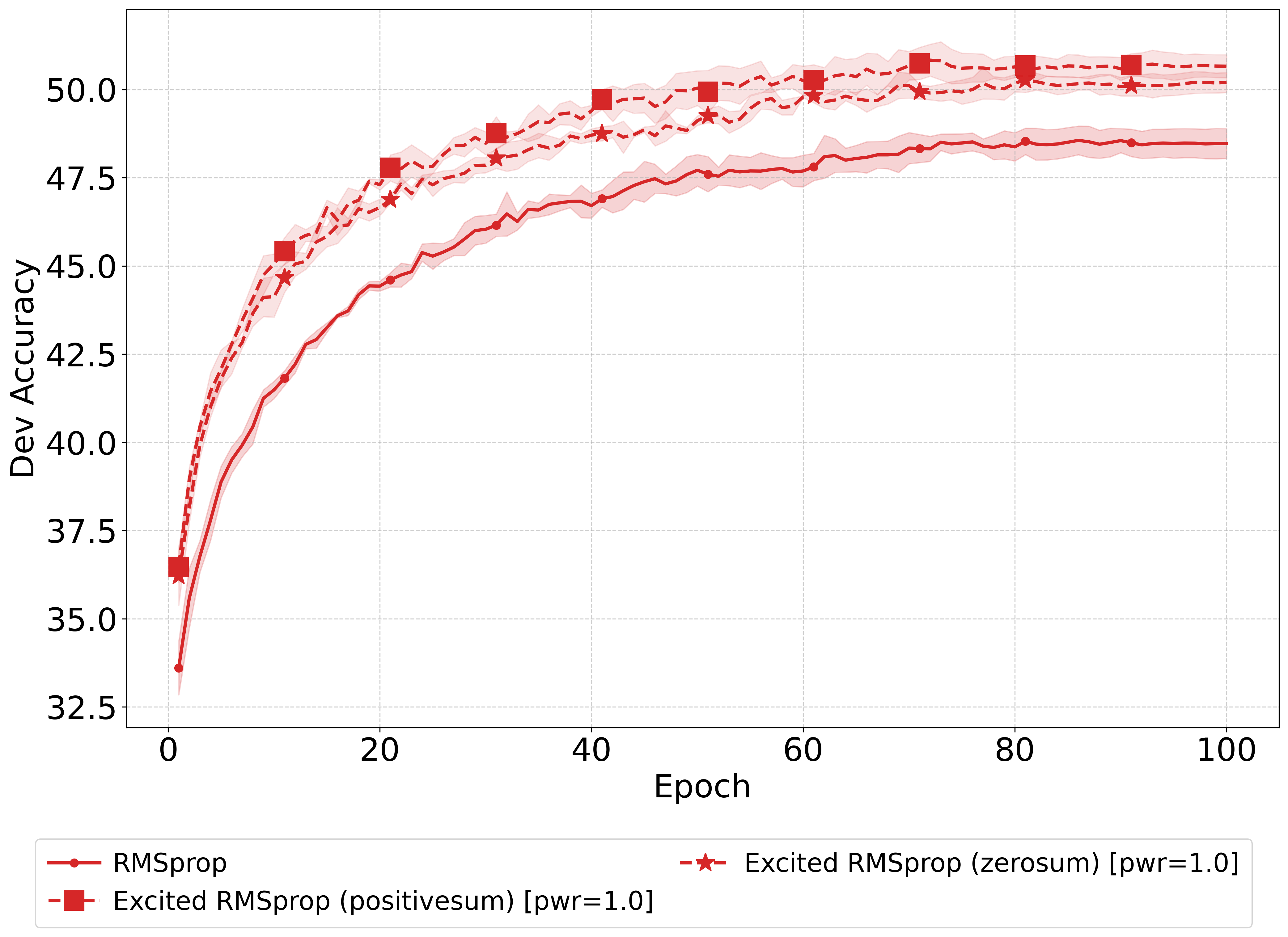}
          \caption{RMSprop}
          \label{fig:cifar10_rmsprop}
     \end{subfigure}

     \vspace{1em}

     \begin{subfigure}[b]{0.48\linewidth}
          \centering
          \includegraphics[width=\linewidth]{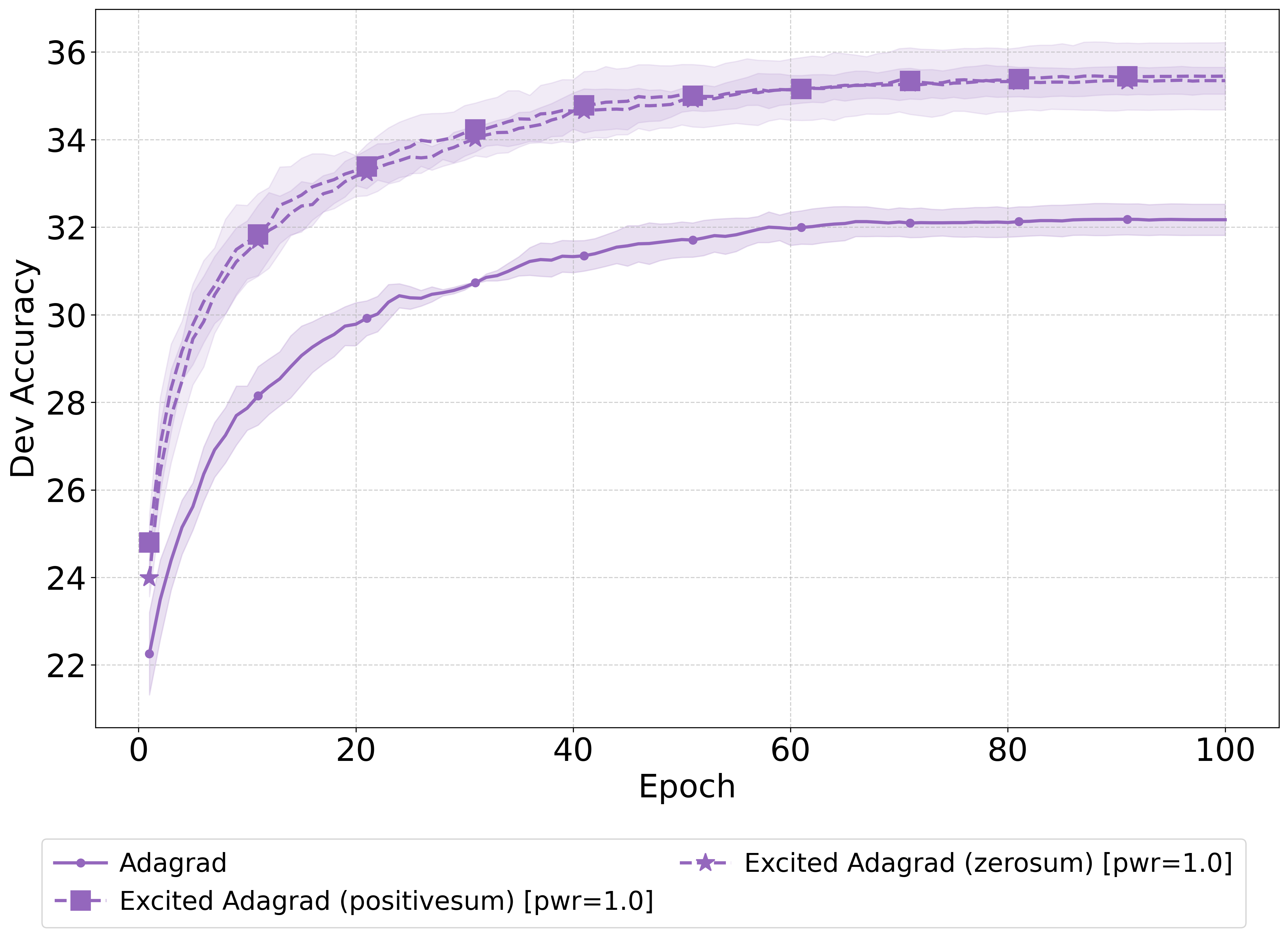}
          \caption{Adagrad}
          \label{fig:cifar10_adagrad}
     \end{subfigure}

     \caption{\textbf{Evolution of Dev Accuracy Across Base Optimizers.} Comparison between vanilla baselines and \textsc{Excitation} variants ($\Phi_{ZS}$, $\Phi_{PS}$) on CIFAR-10. The framework consistently accelerates the ``burn-in'' phase of training and achieves higher terminal stability. The most pronounced gains are observed in first-order and simple adaptive methods like SGD and Adagrad, where the structural signal from \textsc{Excitation} compensates for the lack of complex per-parameter state.}
     \label{fig:all_opt_dynamics_grid}
\end{figure}

As shown in Figure \ref{fig:all_opt_dynamics_grid}, \textsc{Excitation} variants typically enter a regime of superior accuracy within the first 20\% of training iterations. Notably, while the Zero-Sum ($\Phi_{ZS}$) variant often converges faster initially, the Positive-Sum ($\Phi_{PS}$) variant tends to achieve the highest terminal performance, suggesting a trade-off between aggressive expert competition and long-term exploratory stability.

\section{Extended MoE-ViT Results} 
\label{moe_vit_extended}
To verify the generalizability of our architectural findings, we replicate the granularity-vs-capacity experiments on CIFAR-100 and SVHN.

\subsection{CIFAR-100} 
\begin{figure}[ht] \centering \begin{subfigure}[b]{0.48\linewidth} \centering \includegraphics[width=\linewidth]{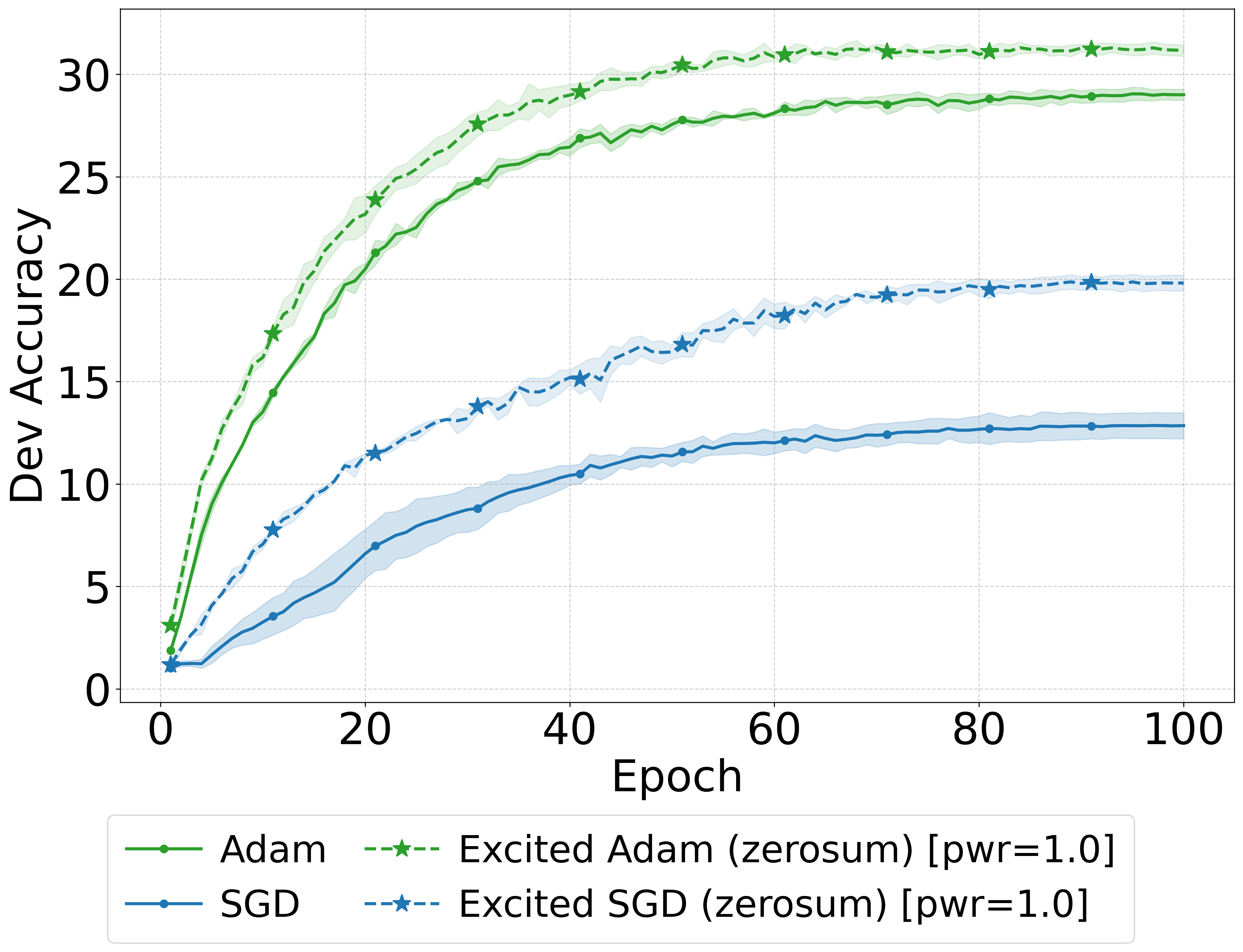} \caption{High Granularity (64 Experts)} \end{subfigure} \hfill \begin{subfigure}[b]{0.48\linewidth} \centering \includegraphics[width=\linewidth]{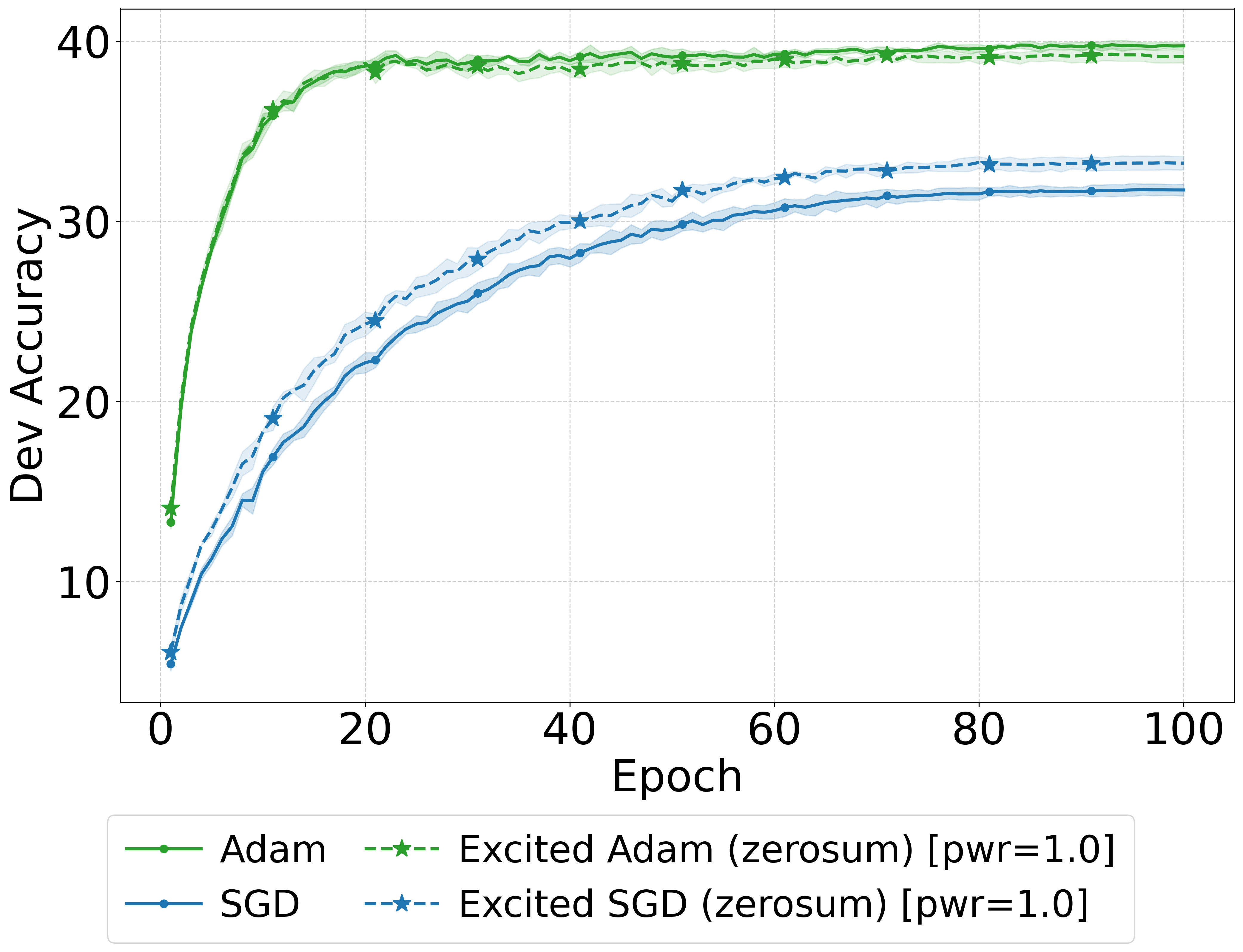} \caption{High Capacity (32 Experts)} \end{subfigure} \caption{\textbf{CIFAR-100 MoE-ViT Scaling.} Results mirror the CIFAR-10 trends, with \textsc{Excitation} providing substantial gains in high-granularity regimes while maintaining parity in high-capacity settings.} \label{fig:app_cifar100} \end{figure}

As shown in Figure \ref{fig:app_cifar100}, the same specialization dynamics persist in the presence of more complex label spaces. In the high-granularity 17M model, \textsc{Excitation} significantly outperforms standard Adam and SGD, mitigating the routing ambiguity that arises when 64 experts compete for a 100-class signal. In the high-capacity 4M model, the framework remains stable, matching the baseline performance and confirming its robustness as a ``safety net.''

\subsection{SVHN}
\label{svhn_moe_vit}
\begin{figure}[ht] \centering \begin{subfigure}[b]{0.48\linewidth} \centering \includegraphics[width=\linewidth]{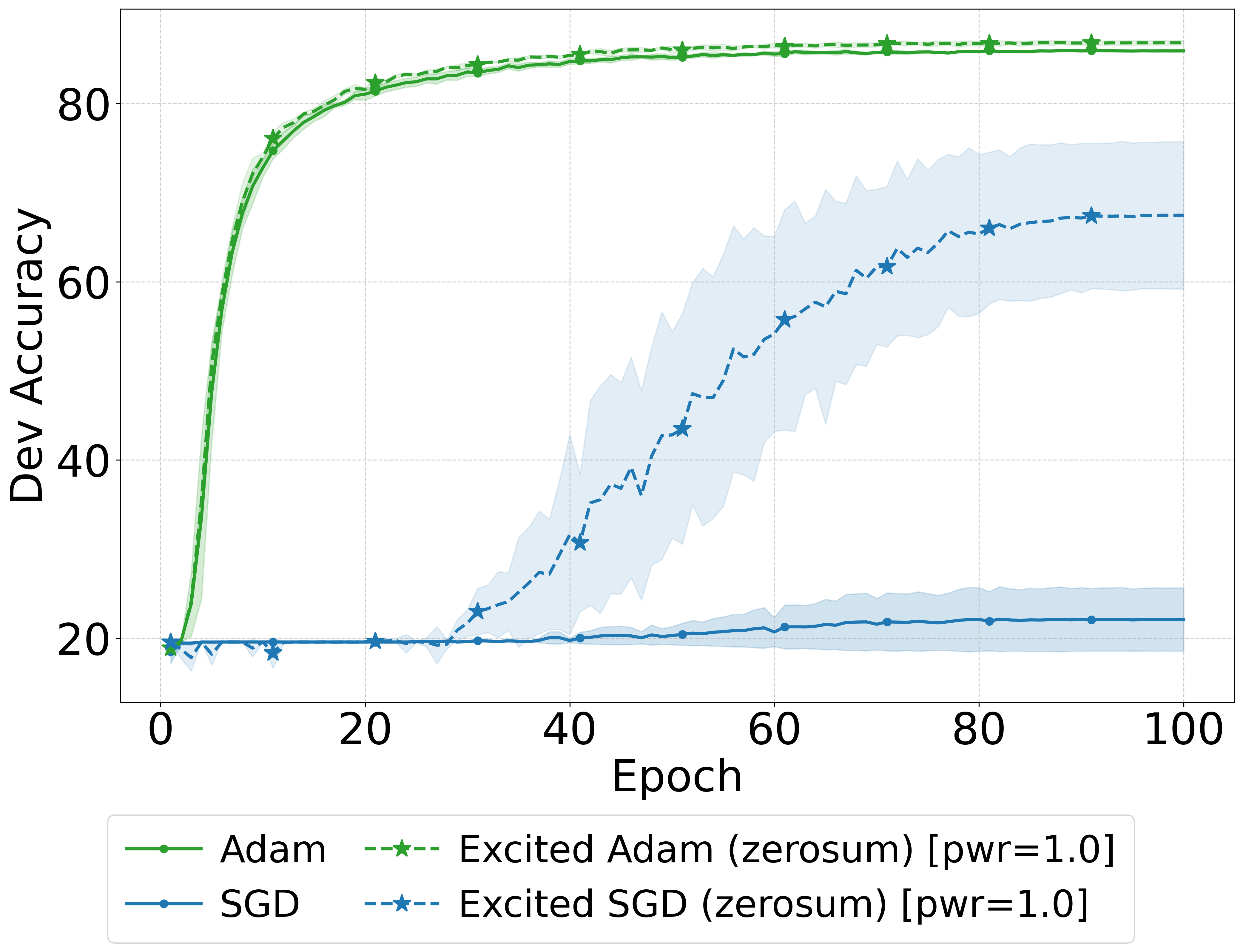} \caption{High Granularity (64 Experts)} \end{subfigure} \hfill \begin{subfigure}[b]{0.48\linewidth} \centering \includegraphics[width=\linewidth]{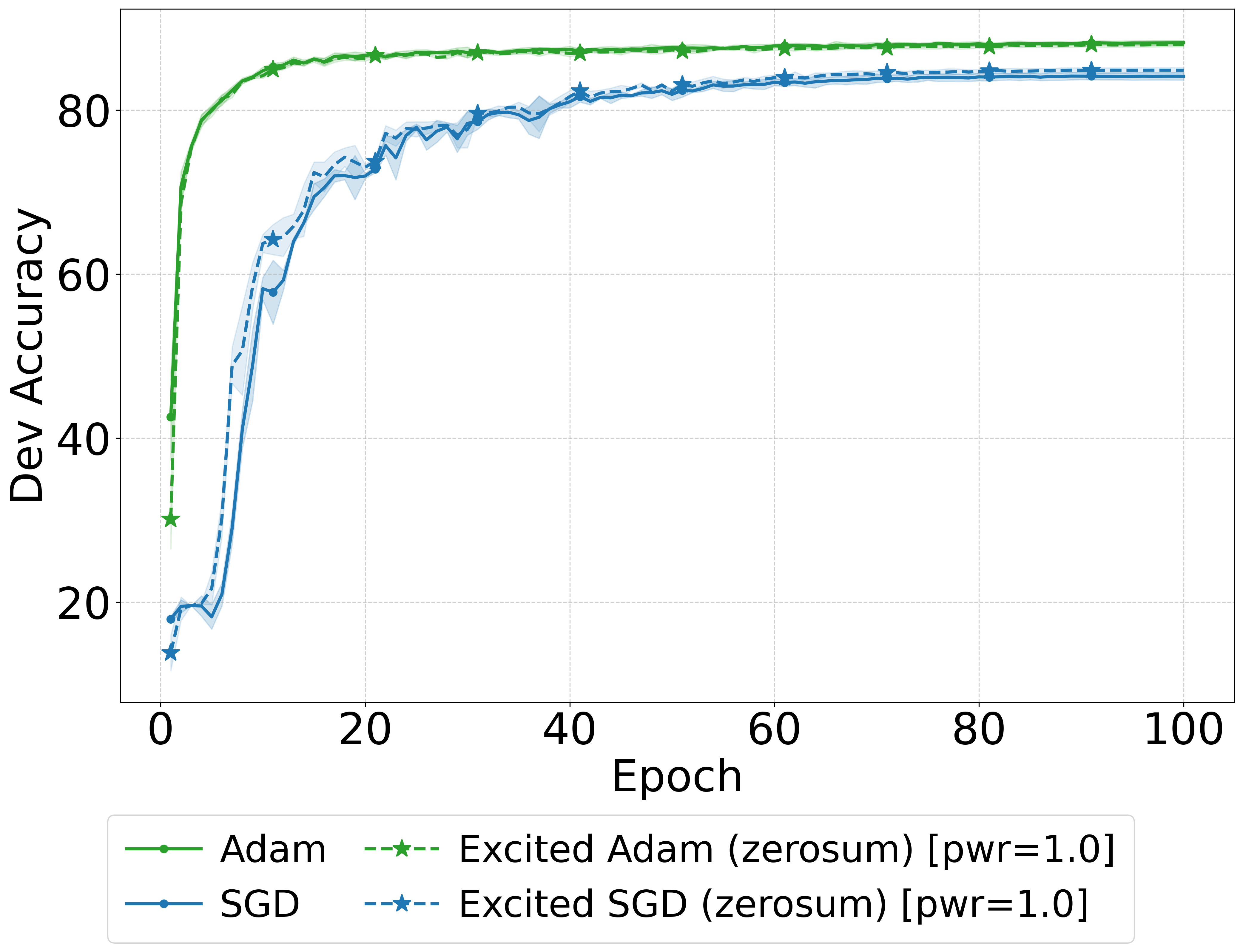} \caption{High Capacity (32 Experts)} \end{subfigure} \caption{\textbf{SVHN MoE-ViT Scaling.} The consistency of performance across datasets suggests that \textsc{Excitation}'s ability to facilitate functional differentiation is a domain-agnostic architectural benefit.} \label{fig:app_svhn} \end{figure}

Similarly, Figure \ref{fig:app_svhn} illustrates these patterns on SVHN. Despite the different data distribution, \textsc{Excitation} consistently rescues the 17M ``tall and skinny'' model from expert redundancy, while remaining passive and safe in the 4M high-utilization regime.

\section{Computational Efficiency}
\label{comp_efficiency}
\begin{table}[h]
\centering
\caption{\textbf{Computational overhead analysis (NVIDIA A100-80GB).} Model configuration defined by width ($W$), layers ($L$), and experts per layer ($E$). $N_e$ denotes total specialized components ($L \times W$ for MLP; $L \times E$ for MoE). We report relative latency overhead $\chi$ ($\pm$ std) across 3 sessions (500 trials each, 100 burn-in). 
}
\label{tab:overhead_analysis}
\small
\setlength{\tabcolsep}{4pt}
\begin{tabular}{@{}llrrS[table-format=2.1, table-figures-uncertainty=1]@{}}
\toprule
\textbf{Architecture} & ($W, L, E$) & \textbf{Params} & \textbf{$N_e$} & \multicolumn{1}{c}{\textbf{Overhead $\chi$ (\%)}} \\ \midrule
\textit{Neuron-level} & 2048, 2, -- & 5.8M & 4,096 & 33.2 \pm 28.4 \\
(Top-$k$ MLP) & 4096, 2, -- & 20.0M & 8,192 & 15.9 \pm 0.1 \\
 & 1024, 16, -- & 16.6M & 16,384 & 29.7 \pm 0.4 \\
 & 16384, 4, -- & 818.4M & 65,536 & 18.7 \pm 0.5 \\ \midrule
\textit{Expert-level} & 512, 6, 8 & 56.8M & 48 & 0.1 \pm 0.1 \\
(Sparse MoE) & 1024, 12, 32 & 1.64B & 384 & -1.2 \pm 1.9 \\ \bottomrule
\end{tabular}
\end{table}
We evaluate the relative overhead $\chi$ across varying routing granularities (Table~\ref{tab:overhead_analysis}). The mechanism incurs a nominal cost of $\mathcal{O}(N_e \cdot B)$ for statistics aggregation and $\mathcal{O}(N_e \cdot d)$ for modulation (where $B$ and $d$ denote batch size and hidden dimension, respectively). This cost is effectively amortized at scale; while measurable in compact MLPs ($33.2\%$), it becomes negligible ($< 0.1\%$) in the 1.64B parameter MoE-ViT. Consequently, \textsc{Excitation} ensures distinct expert partitioning without compromising the throughput benefits of conditional computation.

\end{document}